# Development of an AI Anti-Bullying System Using Large Language Model Key Topic Detection

Matthew Tassava, Cameron Kolodjski, Jordan Milbrath, Adorah Bishop,
Nathan Flanders, Robbie Fetsch, Danielle Hanson, Jeremy Straub

Department of Computer Science
North Dakota State University
1320 Albrecht Blvd., Room 258
Fargo, ND 58108

Phone: +1 (701) 231-8196
Email: jeremy.straub@ndsu.edu (corresponding author)

**Abstract**

This paper presents and evaluates work on the development of an artificial intelligence (AI) anti-bullying system. The system is designed to identify coordinated bullying attacks via social media and other mechanisms, characterize them and propose remediation and response activities to them. In particular, a large language model (LLM) is used to populate an enhanced expert system-based network model of a bullying attack. This facilitates analysis and remediation activity – such as generating report messages to social media companies – determination. The system is described and the efficacy of the LLM for populating the model is analyzed herein.

## 1. Introduction

Cyberbullying involves using technology, such as social media, to mistreat and harass others. It has become a pronounced problem due to the increasing ubiquity of online platforms that provide a means to conduct it. A significant amount of this cyberbullying is conducted by and targets teenagers.

It is difficult for teenage students to shut themselves off from the digital world in which the cyberbullying is taking place. Given how entrenched the use of digital apps is by today's youth, and the pronounced consequences of it – including victim self-harm, in some cases – cyberbullying is at least as much of a threat as physical bullying. Additionally, because of the obfuscation caused by the online environment, authorities (such as parents, teachers and law enforcement) may have difficulty determining what has occurred and who the actors participating are. Determining what can be done to remediate the situation is even more difficult.

Cyberbullying using AI can be particularly pronounced, as AI use provides a wealth of knowledge, incredible speed and other capabilities that can intensify a bullying campaign. This provides significant capabilities to a bully – particularly in regards to youth victims – even when the bully is another youth. For example, the speed of AI bullying could result in the victim feeling overwhelmed and unable to respond fast enough. The knowledge base of a large language model (LLM) creates an asymmetry, where an AI/LLM user may be generating bullying content that neither they – nor the victim – fully understand. This content may be particularly vile and effective. LLMs ability to write content with different tones, from different perspectives and even in different languages can create the impression that numerous people are posting similar content, making it more likely to be believed and intensifying the victim's feeling of being ganged up on.

A partial solution to this challenge is to give responders – parents, teachers, school administrators, counselors, law enforcement and others – a capability that is just as effective as that which a basic LLM provides to a would-be bully. This paper, thus, proposes an AI anti-bullying system (AABS). The system utilizes generative artificial intelligence (GAI) to analyze cyberbullying related messages circulating online and builds a model that clearly represents the transpired events. This model can then be used to generate reports and suggest remediation strategies to necessary authorities, such as teachers, parents, and school administrators. The proposed system is described and evaluated herein.

This paper continues, in Section 2, with a review of prior relevant work.  Then, in Section 3, the proposed technology is described.  Section 4 presents the experimentation and analysis conducted to evaluate the proposed technique.  The technique's prospective impact is discussed in Section 5.  Finally, Section 6 discusses the outcomes of this work and needed areas of future research.

## 2. Background

AI has become the "latest weapon in the arsenal of school bullies" [1].  It is used in cyberbullying to create "deepfakes and other deceitful content" [2], which are highly offensive, to target victims.  It is being used "produce realistic, degrading and offensive content of children" and number of families reporting cyberbullying is growing [3], globally.  Teens are primary targets: the Australian eSafety Commissioner, for example, found that two-thirds of reports of bullying had a victim between 12 and 15 and often continued "bullying occurring on school grounds" [3].  In the United States, 59% of teens reported being bullied online, in one study [4].

Schools and law enforcement are unprepared and, in addition to the direct harm it has caused, AI-based bullying has led to officials taking actions against victims, based on deceptive AI-generated content [5].  The legal system is also struggling to catch up, as there are only a limited number of laws directly covering these activities [5], though they may be proscribed by more general laws.  While new laws are being introduced [6], a recent survey found that over 60% of school district technology leaders are "'very' or 'extremely' concerned" about AI, including its potential use in cyberbullying [7].

This section discusses bullying response and mitigation (in Section 2.1) and technologies developed for bullying detection (in Section 2.2) and response (in Section 2.3).  Finally, it concludes with a discussion regarding the use of AI in cyberbullying prevention in Section 2.4.

### 2.1. Bullying Response and Mitigation

A variety of research has been performed regarding response to bullying and cyberbullying.

Holt, et al. [8] investigated the development of programs and program components to reduce bullying by changing the way teachers and community members respond to bullying. It was found that schools have opportunities to stop bullying, but do not use them.  The community has a significant impact on a victim's ability to get help but needs a way to enforce its anti-bullying policies. Other sources of support for victims include parents, friends, and teachers. It was found that integrating daily measures that promote an anti-bullying environment helps to reduce the number of students that get bullied.

Milosevic, et al. [9] performed qualitative research with pre-teen and teenage children. Their study indicated that AI would help prevent cyber bullying, and some respondents indicated that they would support using AI to prevent cyberbullying. However, most respondents indicated that they would not use AI to prevent cyberbullying because the AI would have to read the messages. Opponents to AI use raised privacy concerns and didn't want face backlash from peers for supporting AI use in cyberbullying prevention.

During the study [9], respondents were presented with several scenarios for AI use.  Most supported AI's use to identify and hide bullying, while only some supported hiding content that had the intention of excluding others. Respondents also supported providing functionality to report bullying content to a school administrator; however, many did not want to report bullying because of peer pressure. Finally, respondents were less supportive of a tool to allow victims to create an anti-bullying video to send it to the bully, as it could simply be ignored or might serve as an indication that the bullying is affecting its victims.

Although bullying rates vary geographically, Saget, Yarmina and Planas [10] found that victims aged 12 to 18 showed a direct correlation with social maladjustments. To combat this, they proposed robotherapy.  They found that the majority of international and national anti-bullying programs (which occurred 20 to 25 years ago) had success in reducing bullying and victimization by improving emotional competency, increasing peer acceptance, involving others, and preventing violence.  Based on their analysis, they proposed a two-component plan that included teenager-robot interaction and psychologist-teenager interaction.  The psychologist taught strategies, such as taking deep breaths, relaxing muscles, maintaining eye contact, and speaking with a calm voice and the robot simulated bullying, and determined whether the child's reaction was "good" or "bad". The study concluded that, although it improved victim response, the approach would be challenging for an AI, as the interactions would have to go beyond just a script.

A variety of other tools can also be used to mitigate and respond to bullying and have shown varying levels of success. Methods – such as social support, social skills, and school-based programs – to reduce the long-term trauma resulting were discussed in [11]. Two European programs were discussed in [12], a long-term anti-bullying program in a Spanish school was discussed in [13], and thirteen anti-bullying programs 2015 were discussed in [14]. A six-session anti-bullying program incorporated into a physical education class was shown to have positive short-term results, in [15], but the program's longer-term impact wasn't studied. Bullying has also been studied in the workplace with Salin, et al. [16], for example, analyzing interview data regarding workplace bullying prevention and response from 214 human resource professionals in fourteen regions. Cortes-Pascual, et al. [17] identified correlations between positive relationships with bullying and being bullied, both on and offline, using information from twenty Aragon high schools.

*2.2. Detection Technology*

A variety of technical approaches can be used to detect cyberbullying. Orru, et al. [18], for example, discussed using multiple forms of analysis – including visual analysis, deepfake detection, keystroke dynamics, verbal analysis, and stress detection – to evaluate the risk of various forms of bullying within classrooms. Similarly, Litty, Jahin and Jesan [19] discussed existing methods for detecting cyberbullying on social media platforms. They discussed the advantages, challenges, and types of deep learning methods used and analyzed case studies of using deep learning for cyberbullying detection on social media platforms.

One key problem with creating models of cyberbullying is the ambiguity associated with cyberbullying scenarios. Training using datasets with ambiguous content can impair AI systems' functionality. Thus, identifying and removing ambiguous comments from a cyberbullying dataset allows AI models to be trained to more accurately identify cyberbullying. Gomez, Sztainberg and Trana [20] investigated two machine learning approaches to separate ambiguous and unambiguous statements and tested them on 19,000 YouTube comments. First, they used single-algorithm consensus filtering, which uses a support vector machine learning algorithm for data filtering. Second, they used multi-algorithm consensus filtering. This uses several naive Bayes, support vector and convolutional neural networks. Amazon's Mechanical Turk was used to supply human workers to determine if a comment was truly ambiguous. It was found that the two proposed classifiers provided a notable improvement in performance over the original datasets

Ramezanian and Niemi [21] attempted to mitigate privacy concerns with AI online message screening for cyberbullying content by designing a methodology where messages are screened in phases. Messages only receive the next phase of scrutiny if flagged in the current phase. This method also classifies users into normal, bully, and victim groups over time. The normal group only gets a small portion of their messages checked.

Dhiman, et al. [22] created a model which can be used to detect fake news that was trained on data from Indian news and fake news sources. Ide and Adewale [23] developed a system, designed to work with popular social media website, that filters online bullying messages at the time when they are sent—preventing the intended recipient from ever seeing them.

*2.3. Response Technology*

Children spend a notable amount of time socializing on mobile applications and, as such, cyberbullying among children takes place on these apps. In response, anti-bullying apps have been developed; however, typically, those in the target demographic of these apps are not involved in their design or development process. Hartikainen, et al. [24] explored involving children aged 7 to 15 in the design and prototyping of anti-bullying apps. Each small group brainstormed ideas and developed a prototype design. Each app had different features including facilitating adult support, peer support, self-help, providing entertainment, and tracking bullying incidents. Students were surveyed at the beginning and throughout the project. Additionally, the views of both students and teachers were analyzed at the project's end.

A variety of other resources have also been developed to help children cope with cyberbullying. For example, a chatbot that was created to provide an anti-bullying resource to students, named CAPTAIN, was proposed in [25]. CAPTAIN can identify content considered to be bullying, answer questions about cyberbullying, and converse with students. Games have also been developed for the purpose of preventing and detecting bullying among children [26]. A design fiction program was proposed by Venta-Olkkonen [27] which asked children to design tools and programs for bullying prevention and response, without regard to current technological limitations.

A game made to detect bullying and the results of its testing, for evaluating exclusionary behavior based on country of origin in Spanish schools, is discussed in [28]. A graphical language that can be used by non-technical professionals to create anti-cyber bullying story-games is discussed in [29]. An application created for students to report bullying in schools is reviewed in [30]. This application allows students to report and describe bullying and directs the reports to the appropriate school officials.

*2.4. AI Technology Use*

Jain [31] notes the significant potential for AI to serve as a "hero" and "silent guardian" against bullying. This can take multiple forms. A phone app called ReThink, for example, uses AI to identify offensive messages and discourage them from being sent [32]. IBM's Watson AI has been used to monitor social media for bullying and warning of potential self-harm [33]. Aston University's Bullstop [34] software integrates with Twitter and prevents "bullies and trolls" from directly messaging potential victims using the software.

Vanderbilt University Medical Center and Florida State University developed a technology that could accurately identify 92% of those about to commit self-harm [4]. Instagram seeks to "proactively detect bullying" in photos, captions and other text and videos [35], while Facebook has deployed a self-harm identification technology [4]. Teenager Gitanjali Rao's Kindly software identifies bullying language in posts, drawing recognition from UNICEF [36].

## 3. Technology

This section discusses the proposed technology. Section 3.1 discusses the technical challenges that this technology is responsive to. Section 3.2 discusses the technologies that the proposed technology builds upon. Then, in Section 3.3, the technology itself is presented. Finally, Section 3.4 discusses the specific system operations that were conducted for the purposes of collecting the data that is analyzed herein.

*3.1. Technical Challenges*

AI provides a wealth of knowledge, incredible speed and other capabilities that could be useful to a bully. This work focuses on preventing bullying at the middle and high school levels; however, this could be applicable to younger students, college students and the general public.

*Speed* – AI can generate content at a much faster rate than a human. This is also a much faster rate than humans can process and respond to content. Thus, an individual being bullied by an AI may feel overwhelmed. They may also be unable to respond to the bullying activities, request post removal and take other needed actions at the speed of the AI.

*Knowledge* – Large language models (LLMs) are trained with a wealth of human knowledge. This far exceeds any one individual's knowledge. LLMs have knowledge that may go well beyond the experience level of middle and high school students. This creates an asymmetry, where the AI/LLM user may be generating bullying content that neither they – nor the victim – fully understand. This content may be particularly vile and effective.

*Other Capabilities* – A bully could prospectively leverage numerous other LLM capabilities, such as its writing capabilities. LLMs have the ability to write in different styles – and even in different languages – and the ability to write at different reading/age levels and with different levels of professionalism. This could be used to create numerous forms of content that prospectively appear to come from different locations.

*3.2. Foundational Technologies*

This technology builds on the Blackboard Architecture, which was introduced by Hayes-Roth in the 1980s [37], based on the Hearsay II system [38] which was developed in the late 1970s. The Blackboard Architecture has found numerous uses including tutoring [39], vehicle control [40] and protein modeling [41]. It is part of a sub-field of machine learning called eXplainable Artificial Intelligence (XAI) which is responsive to the limitations of AI systems' ability to "explain their autonomous decisions to human users" and their lack of understandability by humans [42].

A modern Blackboard Architecture implementation [43] serves as the foundation for this work. It has been extended through the use of previously developed links [44], containers [44], common properties [45] and generic rules [45], which allow bullying networks to be more effectively modeled in this work. The Meta Llama 3.1 Instruct LLM [46] is used extensively, in this system, for natural language processing.

### 3.3. Proposed Technical Solution

A technology has been developed to identify a cyberbullying campaign to facilitate connecting the dots for bullying's first responders, including teachers, school administrators and – in pronounced cases – law enforcement. This is done by leveraging AI technologies.

The AABS system has several key components. It includes a search mechanism, which identifies content for processing, a model development component and a response and remediation component. A high-level depiction of the system is presented in Figure 1, which shows the linear flow of information from the search component to the model development component and, from there, to the response and remediation component. A more detailed workflow for AABS is presented in Figure 2.

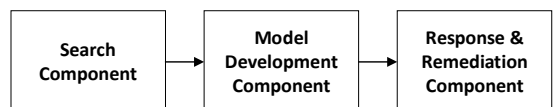

**Figure 1.** AABS high level architecture.

*Search Component*

The search component identifies content for processing by the system. It can make use of interfaces provided by social media providers, search engines and others to identify prospective content that may be indicative of bullying. The filter at this point is very limited. The search component ensures that the content is related to the individual who is prospectively being bullied and within the timeframe that is being studied. If the individual has a common name or other factors result in lots of extraneous content being identified, additional criteria (such as other words to include or words to exclude messages based on) can be supplied. However, the goal is not to try to determine what content may be bullying at this point – it is to cast a proverbial 'wide net' to ensure that all potentially relevant content is considered.

Notably, feedback from the model development component may identify other terms to include or exclude. This can be used to enhance the search component, after initial information is provided by it to the model development component.

*Model Development Component*

The model development component leverages the use of an LLM (specifically, initially using the Llama 3.1 Instruct LLM) to determine whether content is bullying-relevant. If it is, the content is analyzed and added to the bullying model.

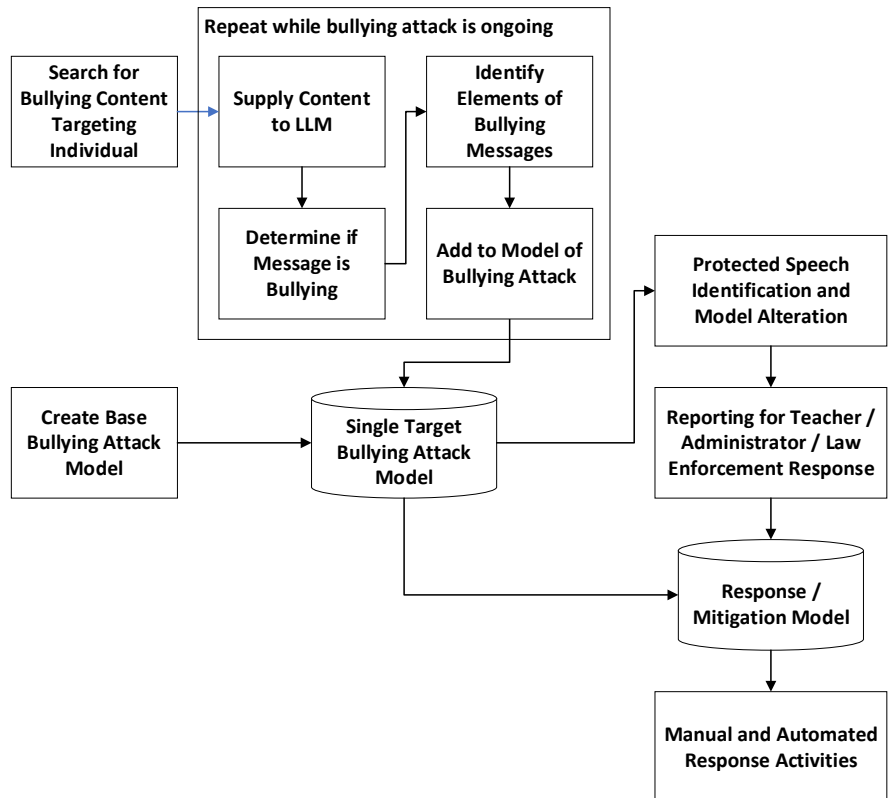

**Figure 2.** AABS workflow overview.

The process starts with the search component supplying the model development component with potentially relevant content to process. The model development component provides this to the LLM, which makes a determination regarding whether bullying is occurring in each message. An example of the type of prompts that can be used for this is provided in Listing 1.

Note that to avoid any impact of LLM protection/guard mechanisms on the answers, this work uses surrogate topics to test the efficacy of LLMs for making this type of determination.

**Listing 1.** Example prompts and responses (responses produced by the Llama 3.1 Instruct model).

> *Prompt:* Determine whether the sentence "Fred went to the market to buy a coat" relates to commerce. Answer only with a single word, either yes or no.
>
> *Response:* yes
>
> *Prompt:* Determine whether the sentence "Fred went to the market to buy a coat" relates to boating. Answer only with a single word, either yes or no.
>
> *Response:* No

Next, the model development component identifies critical areas of the text which serve as the basis for the model being developed. The subject, object, action and other key components of the text are identified. These are used to translate the human-readable text into a basic model. A LLM is used for this processing as well. An example of the type of prompts that can perform this is provided in Listing 2.

**Listing 2.** Example deconstruction prompts and responses (responses from Llama 3.1 Instruct model).

> *Prompt:* Identify the subject of the sentence "Fred went to the market to buy a coat". Answer only with the subject.
>
> *Response:* Fred
>
> *Prompt:* Identify the object of the sentence "Fred went to the market to buy a coat". Answer only with the object.
>
> *Response:* a coat

This information is then supplied to a modified Blackboard Architecture network for processing and analysis. The modern Blackboard Architecture implementation, from [43], serves as the basis of this with the addition of the concepts of links, containers, generic rules and common properties from [44,45].

The combination of all of the data collected by the model development component produces the single target bullying attack model (STBAM). This model is provided to the response and remediation component.

*Response and Remediation Component*

The response and remediation component begins with a process of identifying bullying activities which are protected speech. For the purposes of this assessment, at the moment, speech which correctly identifies its speaker or is statedly anonymous and which is not threatening or defamatory is treated as protected speech. These elements are identified in the model, but not removed.

Protected speech is not actionable, in many instances (particularly by public entity actors); however, it can demonstrate motive and intent and provide context for unprotected speech and conduct. Notably, if the entire model is comprised of protected speech, this process will end with a simple report stating this. Expansion of this system to facilitate expanded definitions of protected speech, to perform additional processing for non-public entities and to allow a target to respond to protected speech are topics for potential future work.

Once the model has been annotated with what is protected speech, it is output in a standardized prose form to be provided to a LLM to summarize and expand upon. This produces a report that can help teachers, administrators and law enforcement (if required) to understand the bullying attack, including its scope and implications.

The STBAM is provided to the remediation component to identify actions that can be taken to respond to and remediate the impact of the bullying attack. Examples of remediations are posting corrective posts, reporting posts and requesting post removal (depending on the policies of the social media sites they are posted on). The remediation component generates requests that can be sent to relevant parties to request actions. The attack report is also provided to the remediation component, so that it can be incorporated into requests, as relevant. Additionally, this content is incorporated in the response / remediation activities report which summarizes the proposed automated and manual response and remediation activities to take.

### 3.4. System Operations for Testing Content Acquisition

Content acquisition is the primary focus of the analysis presented herein. The AABS uses the Meta Llama 3.1 Instruct LLM with customized system prompts. The processing of a given set of input starts with sending it to the LLM with the system prompt shown in Listing 3. An overview of the process is presented in Figure 3.

**Listing 3.** System prompt for determining if the input has multiple independent clauses.

> *"Is the following input multiple independent clauses? Respond with only a Yes or No. Only respond as a JSON object. For example, \"Dillan ate an apple. He thought it tasted good. \" { \"response\": \"Yes\" }."*

If the response is "No", the prompt—which should be a single sentence—is added to the list of sentences to be processed. If the response is "Yes", the prompt is sent to the LLM again with a system prompt for splitting a paragraph into individual sentences, which is shown in Listing 4.

**Listing 4.** System prompt for splitting a paragraph into sentences.

> *"Split this paragraph into distinct sentences. Only respond as a JSON array. For example if given the text \"This is one sentence. This is another sentence\", respond with [ \"This is one sentence.\", \"This is another sentence.\" ]."*

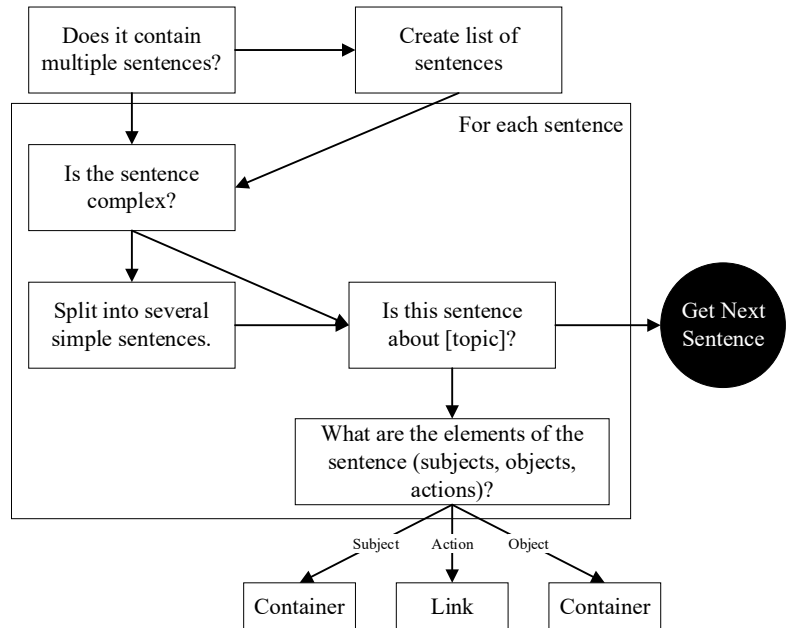

**Figure 3.** System operations overview.

The sentences generated from this prompt are then added to the list of sentences to be processed. Each sentence in the list is processed, one at a time. Processing entails checking whether the sentence is complex or not, by sending it to the LLM with the system prompt shown in Listing 5.

**Listing 5.** System prompt for determining if a sentence needs to be split into multiple simple sentences.

> *"Does this prompt have multiple subjects or objects? If a pronoun relates to a noun in the sentence then it is considered a single subject. Respond with only a Yes or No. Only respond as a JSON object. For example, { \"response\": \"Yes\" }."*

If the response is "Yes", the sentence needs to be split into simple sentences. It is sent to the LLM again with a system prompt, shown in Listing 6, for doing this.

**Listing 6.** System prompt for splitting a sentence into multiple simple sentences.

> *"Split this complex sentence into separate simple sentences made only of a single independent clause. Only respond as a JSON array. For example, given the prompt \"Joe and John bought pencils and markers\" it would give [ \"Joe bought pencils\", \"Joe bought markers\", \"John bought pencils\", \"John bought markers\" ]."*

Each of these sentences is then added to a list of simple sentences to be processed later.

If the initial sentence was not complex (i.e., a response of *"No"* is returned by the complex system prompt), then it is processed right away. To determine if the sentence relates to the key topic, it is passed into the LLM with a topic determination system prompt, shown in Listing 7, where *"{** TOPIC **}"* is replaced with the key topic.

**Listing 7.** System prompt to determine whether a sentence relates to a given topic.

> *"Does the given prompt imply a {** TOPIC **}, whether it is in the past, present, or future? Respond with a Yes or a No. Only respond as a JSON object. For example, { \"response\": \"Yes\" }."*

If the response is "Yes", the sentence is relevant to the key topic, and it needs to be split into its constituent parts. To do this, it is passed into the LLM with a division system prompt, which is shown in Listing 8.

**Listing 8.** System prompt to divide sentences into constituent parts.

*"Analyze the sentence and identify all SVO phrases. Replace any pronouns (e.g. \"he\" or \"she\" or \"they\" or \"I\" or \"you\") with proper nouns. Respond with ONLY a JSON object with only three fields: subject, object, and action. For example, if I say \"Ben is so happy - he bought a boat and is going sailing\", you should reply with { \"subject\": \"Ben\", \"object\": \"a boat\", \"action\": \"bought\" }"*

Each unique subject and object is represented with its own container object. In some cases, the response fields "subject" and "object" may include phrases, such as "car and boat", which must be further split. As such, every subject and object value from the above prompt is passed into the LLM with an element-parsing system prompt. These prompts are shown in Listings 9 and 10, with the first being used to identify subject values and the second being used to identify object values.

**Listing 9.** System prompt to identify subjects.

*"You are given a list of subjects. Identify the subjects. ONLY respond as a JSON array. For example, [ \"Bob\", \"Steve\" ]"*

**Listing 10.** System prompt to identify objects.

*"You are given a list of objects. Identify the objects. ONLY respond as a JSON array. For example, [ \"tape measure\", \"book\" ]"*

The returned objects are then iterated through and turned into container objects. Once the subjects and objects have both been turned into container objects, the link objects are created, as a link needs to reference two containers. The action verb is retrieved from the Listing 8 prompt, and for each combination of subject and object in the sentence, a link is created from the subject to the object with the action verb.

## 4. Experimentation and Results

This section presents the experimentation using and results from testing the proposed technology. Section 4.1 discusses the testing process. Section 4.2 presents the results and discusses their impact.

### 4.1. Testing Process

Due to Llama 3.1's potential for content filtering, testing was not performed on prompts relating to bullying. Data was instead collected on prompts falling into two categories: commerce and travel. The keyword used for first category was "purchase", as other keywords such as "commerce" and "buying or selling" resulted in more erroneous data. The keyword used for the second category was "travel experience", based on the generation prompt used for data creation (which was also performed using the Llama 3.1 Instruct LLM) using the prompts specified in Listings 11 and 12.

**Listing 11.** Test data generation prompt for purchase data.

"Create a list of 10 objects. For each object, write a story about someone buying it."

**Listing 12.** Test data generation prompt for travel experience data.

"Create a list of 10 travel experiences. For each experience, write a story about the experience."

One example from each category (commerce and travel experiences) is presented in this section, in Listings 13 and 14, respectively. All of the input data is presented in Appendix A.

**Listing 13.** Purchase example.

*Prompt:* "Tom had been eyeing a new bike for months, but he finally found the perfect one at his local bike shop. The sleek design and shiny frame caught his eye, and he couldn't wait to take it out for a spin. As he test-rode the bike, he felt a rush of excitement - this was exactly what he had been looking for. The sales associate noticed Tom's enthusiasm and approached him with a smile. "That's one of our top-of-the-line models," he said. "It's designed for speed and comfort, and I think you'll love it." Tom couldn't agree more - he bought the bike on the spot and spent the rest of the day riding through the park."

*Results:* Subjects: [1, Tom]
Objects: [2, bike]
Actions: [1, bought, Tom, bike]

**Listing 14.** Travel experience example.

*Prompt:* "As I soared above the Serengeti National Park in a hot air balloon, I felt a sense of wonder and awe that I had never experienced before. The sun was rising over the savannah, casting a golden glow over the grasslands and acacia trees. Below me, the majestic animals of Africa roamed freely - giraffes, zebras, and lions all seemed to be going about their day as if I wasn't there at all. Our pilot expertly navigated the balloon through the wispy clouds, offering breathtaking views of the landscape below. As we floated above a herd of grazing wildebeest, I felt a sense of peace wash over me - this was truly a once-in-a-lifetime experience."

*Results:* Subjects: [2, The Sun], [3, The Majestic Animals of Africa], [4, The Pilot]
Objects: [1, book], [5, balloon]
Actions: [1, navigated, The Pilot, balloon]

*4.2. Results and Impact*

The results of the network creation activities are presented in Table 1. In this table "expected results" means that the system has correctly modeled at least one example of behavior that is directly associated with the key topic as a link object. For example, if the key topic is "purchase," then actions pertaining to buying and selling would be directly associated, while actions such as a sales associate approaching the buyer would not be directly associated. Note that the data for Test 1 was manually created as an example while the data for all other tests was generated using the previously described technique.

**Table 1.** Evaluation results (with Appendix number listed).

| Example | Got Expected Results | Errors Thrown | # of Subjects Found | # of Objects Found | # of Subjects Used in Actions | # of Objects used in Actions | # of Actions |
|---|---|---|---|---|---|---|---|
| Test 1 (A1) | Yes | No | 1 | 1 | 1 | 1 | 1 |
| Test 2 (A2) | No | No | 0 | 0 | 0 | 0 | 0 |
| Test 3 (A3) | No | No | 0 | 0 | 0 | 0 | 0 |
| Test 4 (A4) | No | Yes | 0 | 0 | 0 | 0 | 0 |
| Test 5 (A5) | No | No | 1 | 1 | 1 | 1 | 1 |
| Test 6 (A6) | No | Yes | 0 | 0 | 0 | 0 | 0 |
| Test 7 (A7) | Yes | No | 1 | 1 | 1 | 1 | 1 |
| Test 8 (A8) | No | Yes | 0 | 0 | 0 | 0 | 0 |
| Test 9 (A9) | No | No | 1 | 0 | 0 | 0 | 0 |
| Test 10 (A10) | No | Yes | 0 | 0 | 0 | 0 | 0 |
| Test 11 (A11) | Yes | No | 1 | 1 | 1 | 1 | 1 |
| Test 12 (A12) | Yes | No | 3 | 2 | 1 | 1 | 1 |
| Test 13 (A13) | Yes | No | 4 | 7 | 4 | 5 | 6 |
| Test 14 (A14) | No | No | 0 | 2 | 0 | 0 | 0 |
| Test 15 (A15) | No | No | 1 | 1 | 0 | 0 | 0 |
| Test 16 (A16) | Yes | No | 2 | 6 | 2 | 2 | 4 |
| Test 17 (A17) | No | No | 2 | 3 | 2 | 2 | 4 |
| Test 18 (A18) | No | No | 3 | 4 | 3 | 1 | 3 |
| Test 19 (A19) | No | No | 3 | 3 | 3 | 1 | 3 |
| Test 20 (A20) | No | No | 1 | 4 | 1 | 1 | 1 |
| Test 21 (A21) | No | No | 2 | 5 | 2 | 1 | 1 |

The errors thrown column indicates whether an error, due to the presence of the term "the entire set" as an object, occurred. The LLM expected an enumerable array of items.

The model creation technology described herein requires additional development to be ready for use outside of the research environment. It was fully successful only 30% of the time. It successfully identified at least part of the subjects, objects and/or actions at least 70% of the time.

With additional enhancement, it could be useful in a variety of fields. First, it could be used to assist in identifying harmful content. It could also find other uses related to cyberbullying. For example, using the proposed technology to analyze victim reports could help to reduce cyberbullying, as it could identify text that could be considered offensive. Then, before any harmful message is sent to another person, certain words could be removed. Alternately, the entire message could be prevented from sending or referred for additional review. This could benefit society by preventing negative messages, improving relationships, helping people communicate effectively, and promoting emotional growth.

Common trends between bullying messages could be discovered, as well. Since the system analyzes text and adds it to a bullying model, a single container that has many links connected to it could serve as an indicator of the relevance of the concept represented by that container. In addition to this, if two containers are highly connected by multiple links, this could imply that the two concepts are closely related. In the context of identifying harmful content, it can be said that two concepts are often used together in the analyzed bullying incidents.

In addition to identifying trends in harmful content, the system has the potential to enhance law enforcement investigations. For instance, in the most extreme cases of school bullying, the process of identification could be accelerated. Typically, school administration is notified by the victim or bystanders, police get called in (if warranted), and the bully faces punishment. Using the proposed technology could greatly speed up this process, since police would be able to gather information more quickly. Reports that contain common information could result in a model that has many links pointing to the container or containers with that information. In reports made from the model, it can be easily determined that many incidents relate to this piece of information, indicating its potential importance in the investigation.

The technology may also have broader security uses. It may be useful to identify data breaches, threats, and malware. Automation, using the technology, may increase efficiency. It may also have application to ensuring peaceful relations with other countries and preventing the spread of terrorist ideologies. It could also be used to detect false information and identify its origins. Being able to relate various reported incidents and stories allows models to be created that represent broad and potentially vague scenarios. The data in the resultant model may provide context and associations between different concepts that would have otherwise gone unnoticed.

**6. Conclusions and Future Work**

The proposed system is not without its flaws. Key topic detection was shown to, currently, be unreliable in a myriad of ways. In some cases, pronouns were not replaced with the proper nouns that were introduced earlier in the user prompt. Some names were be expanded into inferred constituent parts (e.g., "Greek Islands" into Mykonos, Santorini, and Crete). Actions often did not make sense, such as "Petra rising cliffs." These issues can lead to significant cases of unreliability that could negatively impact the system's ability to be useful in certain scenarios and overall trust in the system. However, while it currently may not operate as intended, in some situations, it still has the ability to provide useful associations between concepts that may otherwise go unnoticed. Future development may further enhance the system's efficacy, as well.

There are a number of ways that the system can be improved. The system currently uses Llama 3.1 to process the data. While this model sometimes provides the required functionality for the given test cases, it is possible that an LLM trained specifically to identify key topic-related messages would perform better. This would be particularly important, if the topic might often involve content that would be filtered by the LLM, such as bullying content. In addition to this, editing the system and user prompts that are provided to the LLM could result in more accurate and consistent results. This would require additional testing and could cause the generated models to be more accurate and provide more insightful data.

Further testing of the system is also planned to provide additional data about which user prompts it works well with. Identifying the types of sentences that the system struggles with may allow prompts to be modified to increase the accuracy of results. This may also facilitate supporting input data with greater levels of length and complexity.

A production system will need to incorporate components for searching and analysis, which were not evaluated in this work. Additionally, evidentiary integrity and system cybersecurity will be key considerations.

**Acknowledgements**

Thanks are given to Zach Johnson for helping with elements of this work. This work was supported, in part, by North Dakota EPSCoR.

**Appendix A. Purchase Test Data**

This data, with the exception of the first prompt (which was manually created), was generated using the Llama 3.1 Instruct LLM with the prompt: "Create a list of 10 objects.   For each object, write a story about someone buying it.".

*Prompt:* "Tom had been eyeing a new bike for months, but he finally found the perfect one at his local bike shop. The sleek design and shiny frame caught his eye, and he couldn't wait to take it out for a spin. As he test-rode the bike, he felt a rush of excitement - this was exactly what he had been looking for. The sales associate noticed Tom's enthusiasm and approached him with a smile. "That's one of our top-of-the-line models," he said. "It's designed for speed and comfort, and I think you'll love it." Tom couldn't agree more - he bought the bike on the spot and spent the rest of the day riding through the park."

*Results:*  Subjects: [1, Tom]
         Objects: [2, bike]
         Actions: [1, bought, Tom, bike]

*Prompt:* "My name is Joe. I went into a Target and bought pencils. Then, I bought sushi and Uncrustables to eat for lunch throughout this week."

*Results:*  Subjects: --
         Objects: --
         Actions: --

*Prompt:* "As soon as Emma walked into the music store, she knew she had found her treasure. The vintage guitar on display caught her eye, and she couldn't resist its charm. She carefully picked it up, feeling the weight and balance in her hands. The shop owner noticed her interest and approached her with a smile. "That's a 1960s Fender Stratocaster," he said. "It's been well-maintained and has a rich history." Emma couldn't believe her luck - she had always wanted to play the guitar, and now she had found one that would be perfect for her."

*Results:*  Subjects: --
         Objects: --
         Actions: --

*Prompt:* "Mrs. Thompson was hosting a dinner party for her friends and family, and she wanted everything to be perfect. She walked into the department store and spotted a beautiful set of fine china on display. The delicate patterns and intricate designs caught her eye, and she knew she had found the perfect addition to her table setting. As she carefully examined each piece, the sales associate approached her with a smile. "This is our finest collection," he said. "It's made from high-quality porcelain and will last for generations." Mrs. Thompson couldn't resist - she bought the entire set and spent the rest of the day preparing for her dinner party."

*Results:*  Subjects: --
         Objects: --
         Actions: --
Note: errors out trying to enumerate "the entire set"

*Prompt:* "Professor Patel had been researching ancient civilizations for years, but she was always on the lookout for new insights and perspectives. As she browsed through the bookstore, her eyes landed on a book that caught her attention - "The Lost City of Troy" by a renowned historian. She carefully opened the book and began to read, devouring every word. The author's writing style and in-depth research impressed her, and she knew she had found a valuable resource for her next project. As she purchased the book, the sales associate approached her with a smile. "That's one of our best-sellers," he said. "I think you'll find it fascinating.""

*Results:*  Subjects: [1, The Sales Associate]
         Objects: [2, her]
         Actions: [1, approached, The Sales Associate, her]

*Prompt:* "Lily was an aspiring artist, and she loved nothing more than spending hours in her studio creating new pieces. As she walked into the art supply store, her eyes landed on a set of high-quality paints that caught her attention. She carefully examined each color and texture, feeling inspired by the possibilities. The sales associate noticed Lily's enthusiasm and approached her with a smile. "Those are our top-of-the-line paints," he said. "They're made from the finest materials and will give you incredible results." Lily couldn't resist - she bought the entire set and spent the rest of the day creating new masterpieces."

*Results:* Subjects: -
        Objects: --
        Actions: --
Note: errored out on the entire set

*Prompt:* "Alex had been using his old phone for years, but it was finally time to upgrade. As he walked into the electronics store, his eyes landed on a sleek new smartphone that caught his attention. He carefully examined each feature and function, feeling impressed by its capabilities. The sales associate noticed Alex's interest and approached him with a smile. "That's one of our latest models," he said. "It has all the latest features and is designed for speed and efficiency." Alex couldn't resist - he bought the phone on the spot and spent the rest of the day exploring its new features."

*Results:* Subjects: [1, Alex]
        Objects: [2, phone]
        Actions: [1, bought, Alex, phone]

*Prompt:* "Sarah had always loved gardening, but she was tired of using old, rusty tools that didn't get the job done. As she walked into the garden center, her eyes landed on a set of high-quality gardening tools that caught her attention. She carefully examined each tool and felt impressed by its durability and quality. The sales associate noticed Sarah's enthusiasm and approached her with a smile. "Those are our top-of-the-line tools," he said. "They're made from the finest materials and will make your gardening tasks easier." Sarah couldn't resist - she bought the entire set and spent the rest of the day tending to her garden."

*Results:* Subjects: --
        Objects: --
        Actions: --
Note: entire set error

*Prompt:* "Emily was a student, and she needed a reliable laptop for her studies. As she walked into the electronics store, her eyes landed on a sleek new laptop that caught her attention. She carefully examined each feature and function, feeling impressed by its capabilities. The sales associate noticed Emily's interest and approached her with a smile. "That's one of our latest models," he said. "It has all the latest features and is designed for speed and efficiency." Emily couldn't resist - she bought the laptop on the spot and spent the rest of the day setting it up and exploring its new features."

*Results:* Subjects: [1, Emily]
        Objects: --
        Actions: --

*Prompt:* "Rachel was celebrating her anniversary with her husband, and she wanted to find the perfect gift. As she walked into the jewelry store, her eyes landed on a beautiful set of fine jewelry that caught her attention. She carefully examined each piece and felt impressed by its quality and craftsmanship. The sales associate noticed Rachel's enthusiasm and approached her with a smile. "That's one of our finest collections," he said. "It's made from high-quality materials and will last for generations." Rachel couldn't resist - she bought the entire set and spent the rest of the day planning the perfect anniversary celebration."

*Results:* Subjects: --
        Objects: --

    Actions: --

Note: entire set error

*Prompt:* "Jack had always loved photography, but he was tired of using an old camera that didn't get the job done. As he walked into the electronics store, his eyes landed on a sleek new camera that caught his attention. He carefully examined each feature and function, feeling impressed by its capabilities. The sales associate noticed Jack's enthusiasm and approached him with a smile. "That's one of our top-of-the-line cameras," he said. "It has all the latest features and is designed for speed and efficiency." Jack couldn't resist - he bought the camera on the spot and spent the rest of the day exploring its new features and taking stunning photos."

*Results:* Subjects: [1, Jack]
    Objects: [2, camera]
    Actions: [1, bought, Jack, camera]

## Appendix B. Travel Experience Test Data

This data was created using the Llama 3.1 Instruct LLM with the prompt: "Create a list of 10 travel experiences. For each experience, write a story about the experience.".

*Prompt:* "As I soared above the Serengeti National Park in a hot air balloon, I felt a sense of wonder and awe that I had never experienced before. The sun was rising over the savannah, casting a golden glow over the grasslands and acacia trees. Below me, the majestic animals of Africa roamed freely - giraffes, zebras, and lions all seemed to be going about their day as if I wasn't there at all. Our pilot expertly navigated the balloon through the wispy clouds, offering breathtaking views of the landscape below. As we floated above a herd of grazing wildebeest, I felt a sense of peace wash over me - this was truly a once-in-a-lifetime experience."

*Results:* Subjects: [2, The Sun], [3, The Majestic Animals of Africa], [4, The Pilot]
    Objects: [1, book], [5, balloon]
    Actions: [1, navigated, The Pilot, balloon]

*Prompt:* "As I walked through the narrow canyon known as the Siq, I couldn't help but feel a sense of excitement and wonder. The ancient city of Petra stretched out before me, its intricately carved facades and temples rising up from the sandstone cliffs like giants. I had always been fascinated by history, and to see this incredible site in person was truly a dream come true. As I explored the ruins, I felt a sense of connection to the people who had built this city over 2,000 years ago - their ingenuity and craftsmanship were still evident in every stone."

*Results:* Subjects: [3, Petra], [6, You], [7, The Speaker], [9, I]
    Objects: [1, excitement], [2, wonder], [4, sandstone], [5, cliffs], [8, site], [10, people], [11, city]
    Actions: [1, rising, Petra, sandstone], [2, rising, Petra, cliffs], [3, see, You, site], [4, see, The Speaker, site],
     [5, felt, I, people], [6, felt, I, city]

*Prompt:* "I had always loved Italian food, but nothing could have prepared me for the culinary experience that awaited me in Tuscany. As I walked through the rolling hills and vineyards of this picturesque region, I was invited to join a cooking class at a local villa. Under the guidance of our chef, we learned how to prepare traditional Tuscan dishes using fresh ingredients from the surrounding countryside. The aromas wafting from the kitchen were incredible - pasta sauces simmered on the stovetop, while freshly baked bread and pizza dough rose in the oven. As we sat down to enjoy our creations, I felt a sense of satisfaction and joy that came from sharing a meal with new friends."

*Results:* Subjects: --
    Objects: [1, culinary experience], [2, cooking class]
    Actions: --

*Prompt:* "As I slipped into the crystal-clear waters of Bora Bora's lagoon, I was immediately struck by the vibrant colors and incredible marine life that surrounded me. The sea turtles swam alongside me, their gentle eyes seeming to regard me with a quiet curiosity. As we explored the underwater world together, I felt a sense of connection to these magnificent creatures - they were so much more than just animals; they were ambassadors for the beauty and wonder of our planet."

*Results:* Subjects: [2, Sea Turtles]
Objects: [1, coral reef]
Actions: --

*Prompt:* "As I stood before the majestic Taj Mahal at sunrise, I felt a sense of awe and reverence wash over me. The intricate marble carvings and ornate details seemed to glow with an otherworldly light in the early morning sun. As I walked through the gardens and reflecting pools surrounding this incredible monument, I couldn't help but feel a sense of wonder at the love story that had inspired its creation - Shah Jahan's devotion to his beloved wife was still palpable in every stone."

*Results:* Subjects: [3, marble carvings], [4, ornate details]
Objects: [1, awe], [2, reverence], [5, otherworldly], [6, light], [7, gardens], [8, reflecting pools]
Actions: [1, seemed, marble carvings, otherworldly], [2, seemed, marble carvings, light], [3, seemed, ornate details, otherworldly], [4, seemed, ornate details, light]

*Prompt:* "As I drove through the stunning landscapes of New Zealand's South Island, I felt a sense of freedom and adventure that I had never experienced before. The rugged mountains, pristine lakes, and picturesque villages all seemed to unfold like a tapestry before me - each new turn in the road revealed another breathtaking view or hidden gem. As we stopped at a local café for a cup of coffee and a slice of homemade pie, I felt grateful for this chance to explore one of the most beautiful countries on earth."

*Results:* Subjects: [2, I], [3, me]
Objects: [1, traveler's journal], [4, wall hanging], [5, textile art]
Actions: [1, unfold, I, wall hanging], [2, unfold, I textile art], [3, unfold, me, wall hanging], [4, unfold, me, textile art]

*Prompt:* "As I walked through the crumbling temples and ruins of Angkor Wat, I couldn't help but feel a sense of wonder at the ingenuity and craftsmanship of the people who had built this incredible site over 1,000 years ago. The intricate carvings and bas-reliefs seemed to tell stories of myth and legend, while the imposing stone structures rose up from the jungle like giants. As I explored the ancient city, I felt a sense of connection to the people who had once called this place home - their history and culture still lingered in every stone."

*Results:* Subjects: [3, John], [4, Mary], [5, David]
Objects: [1, temples], [2, ruins], [6, site], [7, city]
Actions: [1, built, John, site], [2, built, Mary, site], [3, built, David, site]

*Prompt:* "As I sailed through the crystal-clear waters of the Aegean Sea, I felt a sense of freedom and joy that I had never experienced before. The picturesque villages and secluded beaches of the Greek Islands seemed to unfold like a treasure map before me - each new cove or hidden harbor revealed another breathtaking view or secret spot. As we anchored in a secluded bay for the night, I felt grateful for this chance to explore one of the most beautiful regions on earth."

*Results:* Subjects: [3, Mykonos], [4, Santorini], [5, Crete]
Objects: [1, freedom], [2, joy], [6, treasure map]
Actions: [1, unfold, Mykonos, treasure map], [2, unfold, Santorini, treasure map], [3, unfold, Crete, treasure map]

*Prompt:* "As I stood before the majestic Grand Canyon at sunset, I felt a sense of awe and wonder wash over me. The rust-red rock formations seemed to glow with an otherworldly light in the fading light of day - the sheer scale and beauty of this incredible natural wonder left me speechless. As I walked along the rim trail, taking in the breathtaking views and listening to the stories of the Native American tribes who had once called this place home, I felt a sense of connection to the land itself."

*Results:* Subjects: [3, The Grand Canyon]
Objects: [1, awe], [2, wonder], [4, you], [5, land]
Actions: [1, left, The Grand Canyon, you]

*Prompt:* "As I rode through the picturesque villages and mountain passes of the Swiss Alps on a vintage train, I felt a sense of nostalgia and wonder wash over me. The snow-capped peaks and turquoise lakes seemed to unfold like a postcard before me - each new station or village revealed another breathtaking view or hidden gem. As we chugged along the tracks, I felt grateful for this chance to experience one of the most scenic train rides in the world."

*Results:* Subjects: [3, mountains], [4, lakes]
Objects: [1, old photograph], [2, childhood memory], [5, postcard], [6, train ride], [7, scenic route]
Actions: [1, seemed, mountains, postcard], [2, seemed, lakes, postcard]

**Appendix C. Experimentation Data**

This data was created using the Llama 3.1 Instruct LLM using the specified prompts in each listing. A workstation with an AMD 7950X3D processor, a MSI NVIDIA 4070 GPU and 32 GB of RAM was used to collect this data.

**Listing A1.** Test 1.
Start: 2:34:31 PM          End: 2:34:55 PM

*System Prompt:* Is the following input multiple independent clauses? Respond with only a Yes or No. Only respond as a JSON object. For example, "Dillan ate an apple. He thought it tasted good. " { "response": "Yes" }.
*Prompt:* Tom had been eyeing a new bike for months, but he finally found the perfect one at his local bike shop. The sleek design and shiny frame caught his eye, and he couldn't wait to take it out for a spin. As he test-rode the bike, he felt a rush of excitement - this was exactly what he had been looking for. The sales associate noticed Tom's enthusiasm and approached him with a smile. "That's one of our top-of-the-line models," he said. "It's designed for speed and comfort, and I think you'll love it." Tom couldn't agree more - he bought the bike on the spot and spent the rest of the day riding through the park.
*Response:* { "response": "Yes" }

*System Prompt:* Split this paragraph into distinct sentences. Only respond as a JSON array For example if given the text "This is one sentence. This is another sentence", respond with [ "This is one sentence.", "This is another sentence." ].
*Prompt:* Tom had been eyeing a new bike for months, but he finally found the perfect one at his local bike shop. The sleek design and shiny frame caught his eye, and he couldn't wait to take it out for a spin. As he test-rode the bike, he felt a rush of excitement - this was exactly what he had been looking for. The sales associate noticed Tom's enthusiasm and approached him with a smile. \"That's one of our top-of-the-line models,\" he said. \"It's designed for speed and comfort, and I think you'll love it.\" Tom couldn't agree more - he bought the bike on the spot and spent the rest of the day riding through the park.
*Response:* [
  "Tom had been eyeing a new bike for months, but he finally found the perfect one at his local bike shop.",
  "The sleek design and shiny frame caught his eye, and he couldn't wait to take it out for a spin.",
  "As he test-rode the bike, he felt a rush of excitement - this was exactly what he had been looking for.",
  "The sales associate noticed Tom's enthusiasm and approached him with a smile.",
  "\"That's one of our top-of-the-line models,\" he said.",
  "\"It's designed for speed and comfort, and I think you'll love it.\"",
  "Tom couldn't agree more - he bought the bike on the spot and spent the rest of the day riding through the park."
]

| | |
|---|---|
| *System Prompt:* | Does this prompt have multiple subjects or objects? If a pronoun relates to a noun in the sentence then it is considered a single subject. Respond with only a Yes or No. Only respond as a JSON object. For example, { "response": "Yes" }. |
| *Prompt:* | Tom had been eyeing a new bike for months, but he finally found the perfect one at his local bike shop. |
| *Response:* | { "response": "No" } |
| | |
| *System Prompt:* | Does the given prompt imply a purchase, whether it is in the past, present, or future? Respond with a Yes or a No. Only respond as a JSON object. For example, { "response": "Yes" }. |
| *Prompt:* | Tom had been eyeing a new bike for months, but he finally found the perfect one at his local bike shop. |
| *Response:* | { "response": "No" } |
| | |
| *System Prompt:* | Does this prompt have multiple subjects or objects? If a pronoun relates to a noun in the sentence then it is considered a single subject. Respond with only a Yes or No. Only respond as a JSON object. For example, { "response": "Yes" }. |
| *Prompt:* | The sleek design and shiny frame caught his eye, and he couldn't wait to take it out for a spin. |
| *Response:* | { "response": "No" } |
| | |
| *System Prompt:* | Does the given prompt imply a purchase, whether it is in the past, present, or future? Respond with a Yes or a No. Only respond as a JSON object. For example, { "response": "Yes" }. |
| *Prompt:* | The sleek design and shiny frame caught his eye, and he couldn't wait to take it out for a spin. |
| *Response:* | { "response": "No" } |
| | |
| *System Prompt:* | Does this prompt have multiple subjects or objects? If a pronoun relates to a noun in the sentence then it is considered a single subject. Respond with only a Yes or No. Only respond as a JSON object. For example, { "response": "Yes" }. |
| *Prompt:* | As he test-rode the bike, he felt a rush of excitement - this was exactly what he had been looking for. |
| *Response:* | { "response": "No" } |
| | |
| *System Prompt:* | Does the given prompt imply a purchase, whether it is in the past, present, or future? Respond with a Yes or a No. Only respond as a JSON object. For example, { "response": "Yes" }. |
| *Prompt:* | As he test-rode the bike, he felt a rush of excitement - this was exactly what he had been looking for. |
| *Response:* | { "response": "No" } |
| | |
| *System Prompt:* | Does this prompt have multiple subjects or objects? If a pronoun relates to a noun in the sentence then it is considered a single subject. Respond with only a Yes or No. Only respond as a JSON object. For example, { "response": "Yes" }. |
| *Prompt:* | The sales associate noticed Tom's enthusiasm and approached him with a smile. |
| *Response:* | { "response": "No" } |
| | |
| *System Prompt:* | Does the given prompt imply a purchase, whether it is in the past, present, or future? Respond with a Yes or a No. Only respond as a JSON object. For example, { "response": "Yes" }. |
| *Prompt:* | The sales associate noticed Tom's enthusiasm and approached him with a smile. |
| *Response:* | { "response": "No" } |
| | |
| *System Prompt:* | Does this prompt have multiple subjects or objects? If a pronoun relates to a noun in the sentence then it is considered a single subject. Respond with only a Yes or No. Only respond as a JSON object. For example, { "response": "Yes" }. |
| *Prompt:* | "That's one of our top-of-the-line models," he said. |
| *Response:* | { "response": "No" } |
| | |
| *System Prompt:* | Does the given prompt imply a purchase, whether it is in the past, present, or future? Respond with a Yes or a No. Only respond as a JSON object. For example, { "response": "Yes" }. |
| *Prompt:* | "That's one of our top-of-the-line models," he said. |
| *Response:* | { "response": "No" } |
| | |
| *System Prompt:* | Does this prompt have multiple subjects or objects? If a pronoun relates to a noun in the sentence then it is considered a single subject. Respond with only a Yes or No. Only respond as a JSON object. For example, { "response": "Yes" }. |
| *Prompt:* | "It's designed for speed and comfort, and I think you'll love it." |
| *Response:* | { "response": "No" } |
| | |
| *System Prompt:* | Does the given prompt imply a purchase, whether it is in the past, present, or future? Respond with a Yes or a No. Only respond as a JSON object. For example, { "response": "Yes" }. |
| *Prompt:* | "It's designed for speed and comfort, and I think you'll love it." |
| *Response:* | { "response": "No" } |
| | |
| *System Prompt:* | Does this prompt have multiple subjects or objects? If a pronoun relates to a noun in the sentence then it is considered a single subject. Respond with only a Yes or No. Only respond as a JSON object. For example, { "response": "Yes" }. |
| *Prompt:* | Tom couldn't agree more - he bought the bike on the spot and spent the rest of the day riding through the park. |
| *Response:* | { "response": "No" } |
| | |
| *System Prompt:* | Does the given prompt imply a purchase, whether it is in the past, present, or future? Respond with a Yes or a No. Only respond as a JSON object. For example, { "response": "Yes" }. |
| *Prompt:* | Tom couldn't agree more - he bought the bike on the spot and spent the rest of the day riding through the park. |
| *Response:* | { "response": "Yes" } |
| | |
| *System Prompt:* | Analyze the sentence and identify all SVO phrases. Replace any pronouns (e.g. "he" or "she" or "they" or "I" or "you") with proper nouns. Respond with ONLY a JSON object with only three fields: subject, object, and action. For example, if I say "Ben is so happy - he bought a boat and is going sailing", you should reply with { "subject": "Ben", "object": "a boat", "action": "bought" } |
| *Prompt:* | Tom couldn't agree more - he bought the bike on the spot and spent the rest of the day riding through the park. |
| *Response:* | {<br>  "subject": "Tom",<br>  "object": "the bike",<br>  "action": "bought"<br>} |

**Listing A2.** Test 2 (manually created sentence prompt).
Start: 2:27:56 PM      End: 2:28:09 PM

| | |
|---|---|
| *System Prompt:* | Is the following input multiple independent clauses? Respond with only a Yes or No. Only respond as a JSON object. For example, "Dillan ate an apple. He thought it tasted good. " { "response": "Yes" }. |
| *Prompt:* | My name is Joe. I went into a Target and bought pencils. Then, I bought sushi and Uncrustables to eat for lunch throughout this week. |
| *Response:* | { "response": "Yes" } |
| | |
| *System Prompt:* | Split this paragraph into distinct sentences. Only respond as a JSON array For example if given the text "This is one sentence. This is another sentence", respond with [ "This is one sentence.", "This is another sentence." ]. |
| *Prompt:* | My name is Joe. I went into a Target and bought pencils. Then, I bought sushi and Uncrustables to eat for lunch throughout this week. |
| *Response:* | [<br>  "My name is Joe.", |

```
            "I went into a Target and bought pencils.",
            "Then, I bought sushi and Uncrustables to eat for lunch throughout this week."
        ]
```

| | |
|---|---|
| *System Prompt:* | Does this prompt have multiple subjects or objects? If a pronoun relates to a noun in the sentence then it is considered a single subject. Respond with only a Yes or No. Only respond as a JSON object. For example, { "response": "Yes" }. |
| *Prompt:* | My name is Joe. |
| *Response:* | { "response": "No" } |
| | |
| *System Prompt:* | Does the given prompt imply a purchase, whether it is in the past, present, or future? Respond with a Yes or a No. Only respond as a JSON object. For example, { "response": "Yes" }. |
| *Prompt:* | My name is Joe. |
| *Response:* | { "response": "No" } |
| | |
| *System Prompt:* | Does this prompt have multiple subjects or objects? If a pronoun relates to a noun in the sentence then it is considered a single subject. Respond with only a Yes or No. Only respond as a JSON object. For example, { "response": "Yes" }. |
| *Prompt:* | I went into a Target and bought pencils. |
| *Response:* | { "response": "No" } |
| | |
| *System Prompt:* | Does the given prompt imply a purchase, whether it is in the past, present, or future? Respond with a Yes or a No. Only respond as a JSON object. For example, { "response": "Yes" }. |
| *Prompt:* | I went into a Target and bought pencils. |
| *Response:* | { "response": "Yes" } |
| | |
| *System Prompt:* | Analyze the sentence and identify all SVO phrases. Replace any pronouns (e.g. "he" or "she" or "they" or "I" or "you") with proper nouns. Respond with ONLY a JSON object with only three fields: subject, object, and action. For example, if I say "Ben is so happy - he bought a boat and is going sailing", you should reply with { "subject": "Ben", "object": "a boat", "action": "bought" } |
| *Prompt:* | I went into a Target and bought pencils. |
| *Response:* | ```
{
  "subject": "you",
  "object": "pencils",
  "action": "bought"
}
``` |
| | |
| *System Prompt:* | Does this prompt have multiple subjects or objects? If a pronoun relates to a noun in the sentence then it is considered a single subject. Respond with only a Yes or No. Only respond as a JSON object. For example, { "response": "Yes" }. |
| *Prompt:* | Then, I bought sushi and Uncrustables to eat for lunch throughout this week. |
| *Response:* | { "response": "No" } |
| | |
| *System Prompt:* | Does the given prompt imply a purchase, whether it is in the past, present, or future? Respond with a Yes or a No. Only respond as a JSON object. For example, { "response": "Yes" }. |
| *Prompt:* | Then, I bought sushi and Uncrustables to eat for lunch throughout this week. |
| *Response:* | { "response": "Yes" } |
| | |
| *System Prompt:* | Analyze the sentence and identify all SVO phrases. Replace any pronouns (e.g. "he" or "she" or "they" or "I" or "you") with proper nouns. Respond with ONLY a JSON object with only three fields: subject, object, and action. For example, if I say "Ben is so happy - he bought a boat and is going sailing", you should reply with { "subject": "Ben", "object": "a boat", "action": "bought" } |
| *Prompt:* | Then, I bought sushi and Uncrustables to eat for lunch throughout this week. |
| *Response:* | ```
{
  "subject": "I",
  "object": "sushi and Uncrustables",
``` |

```
            "action": "bought"
        }
```

**Listing A3.** Test 3.
Start: 2:30:37 PM          End: 2:30:57 PM

*System Prompt:* Is the following input multiple independent clauses? Respond with only a Yes or No. Only respond as a JSON object. For example, "Dillan ate an apple. He thought it tasted good. " { "response": "Yes" }.

*Prompt:* As soon as Emma walked into the music store, she knew she had found her treasure. The vintage guitar on display caught her eye, and she couldn't resist its charm. She carefully picked it up, feeling the weight and balance in her hands. The shop owner noticed her interest and approached her with a smile. "That's a 1960s Fender Stratocaster," he said. "It's been well-maintained and has a rich history." Emma couldn't believe her luck - she had always wanted to play the guitar, and now she had found one that would be perfect for her.

*Response:*
```
{
   "response": "Yes"
}
```

*System Prompt:* Split this paragraph into distinct sentences. Only respond as a JSON array For example if given the text "This is one sentence. This is another sentence", respond with [ "This is one sentence.", "This is another sentence." ].

*Prompt:* As soon as Emma walked into the music store, she knew she had found her treasure. The vintage guitar on display caught her eye, and she couldn't resist its charm. She carefully picked it up, feeling the weight and balance in her hands. The shop owner noticed her interest and approached her with a smile. \"That's a 1960s Fender Stratocaster,\" he said. \"It's been well-maintained and has a rich history.\" Emma couldn't believe her luck - she had always wanted to play the guitar, and now she had found one that would be perfect for her.

*Response:*
```
[
  "As soon as Emma walked into the music store, she knew she had found her treasure.",
  "The vintage guitar on display caught her eye, and she couldn't resist its charm.",
  "She carefully picked it up, feeling the weight and balance in her hands.",
  "The shop owner noticed her interest and approached her with a smile.",
  "\"That's a 1960s Fender Stratocaster,\" he said.",
  "\"It's been well-maintained and has a rich history.\"",
  "Emma couldn't believe her luck - she had always wanted to play the guitar, and now she had found one that would be perfect for her."
]
```

*System Prompt:* Does this prompt have multiple subjects or objects? If a pronoun relates to a noun in the sentence then it is considered a single subject. Respond with only a Yes or No. Only respond as a JSON object. For example, { "response": "Yes" }.
*Prompt:* As soon as Emma walked into the music store, she knew she had found her treasure.
*Response:* { "response": "No" }

*System Prompt:* Does the given prompt imply a purchase, whether it is in the past, present, or future? Respond with a Yes or a No. Only respond as a JSON object. For example, { "response": "Yes" }.
*Prompt:* As soon as Emma walked into the music store, she knew she had found her treasure.
*Response:* { "response": "No" }

*System Prompt:* Does this prompt have multiple subjects or objects? If a pronoun relates to a noun in the sentence then it is considered a single subject. Respond with only a Yes or No. Only respond as a JSON object. For example, { "response": "Yes" }.
*Prompt:* The vintage guitar on display caught her eye, and she couldn't resist its charm.
*Response:* { "response": "No" }

| | |
|---|---|
| *System Prompt:* | Does the given prompt imply a purchase, whether it is in the past, present, or future? Respond with a Yes or a No. Only respond as a JSON object. For example, { "response": "Yes" }. |
| *Prompt:* | The vintage guitar on display caught her eye, and she couldn't resist its charm. |
| *Response:* | { "response": "No" } |
| | |
| *System Prompt:* | Does this prompt have multiple subjects or objects? If a pronoun relates to a noun in the sentence then it is considered a single subject. Respond with only a Yes or No. Only respond as a JSON object. For example, { "response": "Yes" }. |
| *Prompt:* | She carefully picked it up, feeling the weight and balance in her hands. |
| *Response:* | { "response": "No" } |
| | |
| *System Prompt:* | Does the given prompt imply a purchase, whether it is in the past, present, or future? Respond with a Yes or a No. Only respond as a JSON object. For example, { "response": "Yes" }. |
| *Prompt:* | She carefully picked it up, feeling the weight and balance in her hands. |
| *Response:* | { "response": "No" } |
| | |
| *System Prompt:* | Does this prompt have multiple subjects or objects? If a pronoun relates to a noun in the sentence then it is considered a single subject. Respond with only a Yes or No. Only respond as a JSON object. For example, { "response": "Yes" }. |
| *Prompt:* | The shop owner noticed her interest and approached her with a smile. |
| *Response:* | { "response": "No" } |
| | |
| *System Prompt:* | Does the given prompt imply a purchase, whether it is in the past, present, or future? Respond with a Yes or a No. Only respond as a JSON object. For example, { "response": "Yes" }. |
| *Prompt:* | The shop owner noticed her interest and approached her with a smile. |
| *Response:* | { "response": "No" } |
| | |
| *System Prompt:* | Does this prompt have multiple subjects or objects? If a pronoun relates to a noun in the sentence then it is considered a single subject. Respond with only a Yes or No. Only respond as a JSON object. For example, { "response": "Yes" }. |
| *Prompt:* | "That's a 1960s Fender Stratocaster," he said. |
| *Response:* | { "response": "No" } |
| | |
| *System Prompt:* | Does the given prompt imply a purchase, whether it is in the past, present, or future? Respond with a Yes or a No. Only respond as a JSON object. For example, { "response": "Yes" }. |
| *Prompt:* | "That's a 1960s Fender Stratocaster," he said. |
| *Response:* | { "response": "No" } |
| | |
| *System Prompt:* | Does this prompt have multiple subjects or objects? If a pronoun relates to a noun in the sentence then it is considered a single subject. Respond with only a Yes or No. Only respond as a JSON object. For example, { "response": "Yes" }. |
| *Prompt:* | "It's been well-maintained and has a rich history." |
| *Response:* | { "response": "No" } |
| | |
| *System Prompt:* | Does the given prompt imply a purchase, whether it is in the past, present, or future? Respond with a Yes or a No. Only respond as a JSON object. For example, { "response": "Yes" }. |
| *Prompt:* | "It's been well-maintained and has a rich history." |
| *Response:* | { "response": "No" } |
| | |
| *System Prompt:* | Does this prompt have multiple subjects or objects? If a pronoun relates to a noun in the sentence then it is considered a single subject. Respond with only a Yes or No. Only respond as a JSON object. For example, { "response": "Yes" }. |
| *Prompt:* | Emma couldn't believe her luck - she had always wanted to play the guitar, and now she had found one that would be perfect for her. |
| *Response:* | { "response": "No" } |
| *System Prompt:* | Does the given prompt imply a purchase, whether it is in the past, present, or future? Respond with a Yes or a No. Only respond as a JSON object. For example, { "response": "Yes" }. |
| *Prompt:* | Emma couldn't believe her luck - she had always wanted to play the guitar, and now she had found one that would be perfect for her. |
| *Response:* | { "response": "No" } |

**Listing A4.** Test 4.
Start: 2:31:26 PM          End: Didn't Complete - Error

| | |
|---|---|
| *System Prompt:* | Is the following input multiple independent clauses? Respond with only a Yes or No. Only respond as a JSON object. For example, "Dillan ate an apple. He thought it tasted good. " { "response": "Yes" }. |
| *Prompt:* | Mrs. Thompson was hosting a dinner party for her friends and family, and she wanted everything to be perfect. She walked into the department store and spotted a beautiful set of fine china on display. The delicate patterns and intricate designs caught her eye, and she knew she had found the perfect addition to her table setting. As she carefully examined each piece, the sales associate approached her with a smile. "This is our finest collection," he said. "It's made from high-quality porcelain and will last for generations." Mrs. Thompson couldn't resist - she bought the entire set and spent the rest of the day preparing for her dinner party. |
| *Response:* | {<br>  "response": "Yes"<br>} |
| *System Prompt:* | Split this paragraph into distinct sentences. Only respond as a JSON array For example if given the text "This is one sentence. This is another sentence", respond with [ "This is one sentence.", "This is another sentence." ]. |
| *Prompt:* | Mrs. Thompson was hosting a dinner party for her friends and family, and she wanted everything to be perfect. She walked into the department store and spotted a beautiful set of fine china on display. The delicate patterns and intricate designs caught her eye, and she knew she had found the perfect addition to her table setting. As she carefully examined each piece, the sales associate approached her with a smile. \"This is our finest collection,\" he said. \"It's made from high-quality porcelain and will last for generations.\" Mrs. Thompson couldn't resist - she bought the entire set and spent the rest of the day preparing for her dinner party. |
| *Response:* | [<br>  "Mrs. Thompson was hosting a dinner party for her friends and family, and she wanted everything to be perfect.",<br>  "She walked into the department store and spotted a beautiful set of fine china on display.",<br>  "The delicate patterns and intricate designs caught her eye, and she knew she had found the perfect addition to her table setting.",<br>  "As she carefully examined each piece, the sales associate approached her with a smile.",<br>  "\"This is our finest collection,\" he said.",<br>  "\"It's made from high-quality porcelain and will last for generations.\"",<br>  "Mrs. Thompson couldn't resist - she bought the entire set and spent the rest of the day preparing for her dinner party."<br>] |
| *System Prompt:* | Does this prompt have multiple subjects or objects? If a pronoun relates to a noun in the sentence then it is considered a single subject. Respond with only a Yes or No. Only respond as a JSON object. For example, { "response": "Yes" }. |
| *Prompt:* | Mrs. Thompson was hosting a dinner party for her friends and family, and she wanted everything to be perfect. |
| *Response:* | { "response": "No" } |
| *System Prompt:* | Does the given prompt imply a purchase, whether it is in the past, present, or future? Respond with a Yes or a No. Only respond as a JSON object. For example, { "response": "Yes" }. |
| *Prompt:* | Mrs. Thompson was hosting a dinner party for her friends and family, and she wanted everything |
| | to be perfect. |
| *Response:* | { "response": "No" } |
| | |
| *System Prompt:* | Does this prompt have multiple subjects or objects? If a pronoun relates to a noun in the sentence then it is considered a single subject. Respond with only a Yes or No. Only respond as a JSON object. For example, { "response": "Yes" }. |
| *Prompt:* | She walked into the department store and spotted a beautiful set of fine china on display. |
| *Response:* | { "response": "No" } |
| | |
| *System Prompt:* | Does the given prompt imply a purchase, whether it is in the past, present, or future? Respond with a Yes or a No. Only respond as a JSON object. For example, { "response": "Yes" }. |
| *Prompt:* | She walked into the department store and spotted a beautiful set of fine china on display. |
| *Response:* | { "response": "No" } |
| | |
| *System Prompt:* | Does this prompt have multiple subjects or objects? If a pronoun relates to a noun in the sentence then it is considered a single subject. Respond with only a Yes or No. Only respond as a JSON object. For example, { "response": "Yes" }. |
| *Prompt:* | The delicate patterns and intricate designs caught her eye, and she knew she had found the perfect addition to her table setting. |
| *Response:* | { "response": "No" } |
| | |
| *System Prompt:* | Does the given prompt imply a purchase, whether it is in the past, present, or future? Respond with a Yes or a No. Only respond as a JSON object. For example, { "response": "Yes" }. |
| *Prompt:* | The delicate patterns and intricate designs caught her eye, and she knew she had found the perfect addition to her table setting. |
| *Response:* | { "response": "No" } |
| | |
| *System Prompt:* | Does this prompt have multiple subjects or objects? If a pronoun relates to a noun in the sentence then it is considered a single subject. Respond with only a Yes or No. Only respond as a JSON object. For example, { "response": "Yes" }. |
| *Prompt:* | As she carefully examined each piece, the sales associate approached her with a smile. |
| *Response:* | { "response": "No" } |
| | |
| *System Prompt:* | Does the given prompt imply a purchase, whether it is in the past, present, or future? Respond with a Yes or a No. Only respond as a JSON object. For example, { "response": "Yes" }. |
| *Prompt:* | As she carefully examined each piece, the sales associate approached her with a smile. |
| *Response:* | { "response": "No" } |
| | |
| *System Prompt:* | Does this prompt have multiple subjects or objects? If a pronoun relates to a noun in the sentence then it is considered a single subject. Respond with only a Yes or No. Only respond as a JSON object. For example, { "response": "Yes" }. |
| *Prompt:* | "This is our finest collection," he said. |
| *Response:* | { "response": "No" } |
| | |
| *System Prompt:* | Does the given prompt imply a purchase, whether it is in the past, present, or future? Respond with a Yes or a No. Only respond as a JSON object. For example, { "response": "Yes" }. |
| *Prompt:* | "This is our finest collection," he said. |
| *Response:* | { "response": "No" } |
| | |
| *System Prompt:* | Does this prompt have multiple subjects or objects? If a pronoun relates to a noun in the sentence then it is considered a single subject. Respond with only a Yes or No. Only respond as a JSON object. For example, { "response": "Yes" }. |
| *Prompt:* | "It's made from high-quality porcelain and will last for generations." |
| *Response:* | { "response": "No" } |
| | |
| *System Prompt:* | Does the given prompt imply a purchase, whether it is in the past, present, or future? Respond |
| | with a Yes or a No. Only respond as a JSON object. For example, { "response": "Yes" }. |
| *Prompt:* | "It's made from high-quality porcelain and will last for generations." |
| *Response:* | { "response": "No" } |
| *System Prompt:* | Does this prompt have multiple subjects or objects? If a pronoun relates to a noun in the sentence then it is considered a single subject. Respond with only a Yes or No. Only respond as a JSON object. For example, { "response": "Yes" }. |
| *Prompt:* | Mrs. Thompson couldn't resist - she bought the entire set and spent the rest of the day preparing for her dinner party. |
| *Response:* | { "response": "No" } |
| *System Prompt:* | Does the given prompt imply a purchase, whether it is in the past, present, or future? Respond with a Yes or a No. Only respond as a JSON object. For example, { "response": "Yes" }. |
| *Prompt:* | Mrs. Thompson couldn't resist - she bought the entire set and spent the rest of the day preparing for her dinner party. |
| *Response:* | { "response": "Yes" } |
| *System Prompt:* | Analyze the sentence and identify all SVO phrases. Replace any pronouns (e.g. "he" or "she" or "they" or "I" or "you") with proper nouns. Respond with ONLY a JSON object with only three fields: subject, object, and action. For example, if I say "Ben is so happy - he bought a boat and is going sailing", you should reply with { "subject": "Ben", "object": "a boat", "action": "bought" } |
| *Prompt:* | Mrs. Thompson couldn't resist - she bought the entire set and spent the rest of the day preparing for her dinner party. |
| *Response:* | {<br>  "subject": "Mrs. Thompson",<br>  "object": "the entire set",<br>  "action": "bought"<br>} |

**Listing A5.** Test 5.
Start: 2:36:19 PM          End: 2:36:41 PM

|  |  |
|---|---|
| *System Prompt:* | Is the following input multiple independent clauses? Respond with only a Yes or No. Only respond as a JSON object. For example, "Dillan ate an apple. He thought it tasted good. " { "response": "Yes" }. |
| *Prompt:* | Professor Patel had been researching ancient civilizations for years, but she was always on the lookout for new insights and perspectives. As she browsed through the bookstore, her eyes landed on a book that caught her attention - "The Lost City of Troy" by a renowned historian. She carefully opened the book and began to read, devouring every word. The author's writing style and in-depth research impressed her, and she knew she had found a valuable resource for her next project. As she purchased the book, the sales associate approached her with a smile. "That's one of our best-sellers," he said. "I think you'll find it fascinating. |
| *Response:* | { "response": "Yes" } |
| *System Prompt:* | Split this paragraph into distinct sentences. Only respond as a JSON array For example if given the text "This is one sentence. This is another sentence", respond with [ "This is one sentence.", "This is another sentence." ]. |
| *Prompt:* | Professor Patel had been researching ancient civilizations for years, but she was always on the lookout for new insights and perspectives. As she browsed through the bookstore, her eyes landed on a book that caught her attention - \"The Lost City of Troy\" by a renowned historian. She carefully opened the book and began to read, devouring every word. The author's writing style and in-depth research impressed her, and she knew she had found a valuable resource for her next project. As she purchased the book, the sales associate approached her with a smile. \"That's one of our best-sellers,\" he said. \"I think you'll find it fascinating. |
| *Response:* | [<br>  "Professor Patel had been researching ancient civilizations for years, but she was always on the |

```
      lookout for new insights and perspectives.",
      "As she browsed through the bookstore, her eyes landed on a book that caught her attention -
\"The Lost City of Troy\" by a renowned historian.",
      "She carefully opened the book and began to read, devouring every word.",
      "The author's writing style and in-depth research impressed her, and she knew she had found a
valuable resource for her next project.",
      "As she purchased the book, the sales associate approached her with a smile.",
      "\"That's one of our best-sellers,\" he said.",
      "\"I think you'll find it fascinating.\""
   ]
```

| | |
|---|---|
| *System Prompt:* | Does this prompt have multiple subjects or objects? If a pronoun relates to a noun in the sentence then it is considered a single subject. Respond with only a Yes or No. Only respond as a JSON object. For example, { "response": "Yes" }. |
| *Prompt:* | Professor Patel had been researching ancient civilizations for years, but she was always on the lookout for new insights and perspectives. |
| *Response:* | { "response": "No" } |
| *System Prompt:* | Does the given prompt imply a purchase, whether it is in the past, present, or future? Respond with a Yes or a No. Only respond as a JSON object. For example, { "response": "Yes" }. |
| *Prompt:* | Professor Patel had been researching ancient civilizations for years, but she was always on the lookout for new insights and perspectives. |
| *Response:* | { "response": "No" } |
| *System Prompt:* | Does this prompt have multiple subjects or objects? If a pronoun relates to a noun in the sentence then it is considered a single subject. Respond with only a Yes or No. Only respond as a JSON object. For example, { "response": "Yes" }. |
| *Prompt:* | As she browsed through the bookstore, her eyes landed on a book that caught her attention - "The Lost City of Troy" by a renowned historian. |
| *Response:* | { "response": "No" } |
| *System Prompt:* | Does the given prompt imply a purchase, whether it is in the past, present, or future? Respond with a Yes or a No. Only respond as a JSON object. For example, { "response": "Yes" }. |
| *Prompt:* | As she browsed through the bookstore, her eyes landed on a book that caught her attention - "The Lost City of Troy" by a renowned historian. |
| *Response:* | { "response": "No" } |
| *System Prompt:* | Does this prompt have multiple subjects or objects? If a pronoun relates to a noun in the sentence then it is considered a single subject. Respond with only a Yes or No. Only respond as a JSON object. For example, { "response": "Yes" }. |
| *Prompt:* | She carefully opened the book and began to read, devouring every word. |
| *Response:* | { "response": "No" } |
| *System Prompt:* | Does the given prompt imply a purchase, whether it is in the past, present, or future? Respond with a Yes or a No. Only respond as a JSON object. For example, { "response": "Yes" }. |
| *Prompt:* | She carefully opened the book and began to read, devouring every word. |
| *Response:* | { "response": "No" } |
| *System Prompt:* | Does this prompt have multiple subjects or objects? If a pronoun relates to a noun in the sentence then it is considered a single subject. Respond with only a Yes or No. Only respond as a JSON object. For example, { "response": "Yes" }. |
| *Prompt:* | The author's writing style and in-depth research impressed her, and she knew she had found a valuable resource for her next project. |
| *Response:* | { "response": "No" } |
| *System Prompt:* | Does the given prompt imply a purchase, whether it is in the past, present, or future? Respond |
| | with a Yes or a No. Only respond as a JSON object. For example, { "response": "Yes" }. |
| *Prompt:* | The author's writing style and in-depth research impressed her, and she knew she had found a valuable resource for her next project. |
| *Response:* | { "response": "No" } |
| | |
| *System Prompt:* | Does this prompt have multiple subjects or objects? If a pronoun relates to a noun in the sentence then it is considered a single subject. Respond with only a Yes or No. Only respond as a JSON object. For example, { "response": "Yes" }. |
| *Prompt:* | As she purchased the book, the sales associate approached her with a smile. |
| *Response:* | { "response": "No" } |
| | |
| *System Prompt:* | Does the given prompt imply a purchase, whether it is in the past, present, or future? Respond with a Yes or a No. Only respond as a JSON object. For example, { "response": "Yes" }. |
| *Prompt:* | As she purchased the book, the sales associate approached her with a smile. |
| *Response:* | { "response": "Yes" } |
| | |
| *System Prompt:* | Analyze the sentence and identify all SVO phrases. Replace any pronouns (e.g. "he" or "she" or "they" or "I" or "you") with proper nouns. Respond with ONLY a JSON object with only three fields: subject, object, and action. For example, if I say "Ben is so happy - he bought a boat and is going sailing", you should reply with { "subject": "Ben", "object": "a boat", "action": "bought" } |
| *Prompt:* | As she purchased the book, the sales associate approached her with a smile. |
| *Response:* | { <br>   "subject": "the sales associate", <br>   "object": "her", <br>   "action": "approached" <br> } |
| | |
| *System Prompt:* | Does this prompt have multiple subjects or objects? If a pronoun relates to a noun in the sentence then it is considered a single subject. Respond with only a Yes or No. Only respond as a JSON object. For example, { "response": "Yes" }. |
| *Prompt:* | "That's one of our best-sellers," he said. |
| *Response:* | { "response": "No" } |
| | |
| *System Prompt:* | Does the given prompt imply a purchase, whether it is in the past, present, or future? Respond with a Yes or a No. Only respond as a JSON object. For example, { "response": "Yes" }. |
| *Prompt:* | "That's one of our best-sellers," he said. |
| *Response:* | { "response": "No" } |
| | |
| *System Prompt:* | Does this prompt have multiple subjects or objects? If a pronoun relates to a noun in the sentence then it is considered a single subject. Respond with only a Yes or No. Only respond as a JSON object. For example, { "response": "Yes" }. |
| *Prompt:* | "I think you'll find it fascinating." |
| *Response:* | { "response": "No" } |
| | |
| *System Prompt:* | Does the given prompt imply a purchase, whether it is in the past, present, or future? Respond with a Yes or a No. Only respond as a JSON object. For example, { "response": "Yes" }. |
| *Prompt:* | "I think you'll find it fascinating." |
| *Response:* | { "response": "No" } |

**Listing A6.** Test 6.
Start: 2:38:00 PM                End: Didn't Complete - Error

| | |
|---|---|
| *System Prompt:* | Is the following input multiple independent clauses? Respond with only a Yes or No. Only respond as a JSON object. For example, "Dillan ate an apple. He thought it tasted good. " { "response": "Yes" }. |
| *Prompt:* | Lily was an aspiring artist, and she loved nothing more than spending hours in her studio creating |
| | new pieces. As she walked into the art supply store, her eyes landed on a set of high-quality paints that caught her attention. She carefully examined each color and texture, feeling inspired by the possibilities. The sales associate noticed Lily's enthusiasm and approached her with a smile. "Those are our top-of-the-line paints," he said. "They're made from the finest materials and will give you incredible results." Lily couldn't resist - she bought the entire set and spent the rest of the day creating new masterpieces. |
| *Response:* | { "response": "Yes" } |
| | |
| *System Prompt:* | Split this paragraph into distinct sentences. Only respond as a JSON array For example if given the text "This is one sentence. This is another sentence", respond with [ "This is one sentence.", "This is another sentence." ]. |
| *Prompt:* | Lily was an aspiring artist, and she loved nothing more than spending hours in her studio creating new pieces. As she walked into the art supply store, her eyes landed on a set of high-quality paints that caught her attention. She carefully examined each color and texture, feeling inspired by the possibilities. The sales associate noticed Lily's enthusiasm and approached her with a smile. \"Those are our top-of-the-line paints,\" he said. \"They're made from the finest materials and will give you incredible results.\" Lily couldn't resist - she bought the entire set and spent the rest of the day creating new masterpieces. |
| *Response:* | [<br>  "Lily was an aspiring artist, and she loved nothing more than spending hours in her studio creating new pieces.",<br>  "As she walked into the art supply store, her eyes landed on a set of high-quality paints that caught her attention.",<br>  "She carefully examined each color and texture, feeling inspired by the possibilities.",<br>  "The sales associate noticed Lily's enthusiasm and approached her with a smile.",<br>  "\"Those are our top-of-the-line paints,\" he said.",<br>  "\"They're made from the finest materials and will give you incredible results.\"",<br>  "Lily couldn't resist - she bought the entire set and spent the rest of the day creating new masterpieces."<br>] |
| | |
| *System Prompt:* | Does this prompt have multiple subjects or objects? If a pronoun relates to a noun in the sentence then it is considered a single subject. Respond with only a Yes or No. Only respond as a JSON object. For example, { "response": "Yes" }. |
| *Prompt:* | Lily was an aspiring artist, and she loved nothing more than spending hours in her studio creating new pieces. |
| *Response:* | { "response": "No" } |
| | |
| *System Prompt:* | Does the given prompt imply a purchase, whether it is in the past, present, or future? Respond with a Yes or a No. Only respond as a JSON object. For example, { "response": "Yes" }. |
| *Prompt:* | Lily was an aspiring artist, and she loved nothing more than spending hours in her studio creating new pieces. |
| *Response:* | { "response": "No" } |
| | |
| *System Prompt:* | Does this prompt have multiple subjects or objects? If a pronoun relates to a noun in the sentence then it is considered a single subject. Respond with only a Yes or No. Only respond as a JSON object. For example, { "response": "Yes" }. |
| *Prompt:* | As she walked into the art supply store, her eyes landed on a set of high-quality paints that caught her attention. |
| *Response:* | { "response": "No" } |
| | |
| *System Prompt:* | Does the given prompt imply a purchase, whether it is in the past, present, or future? Respond with a Yes or a No. Only respond as a JSON object. For example, { "response": "Yes" }. |
| *Prompt:* | As she walked into the art supply store, her eyes landed on a set of high-quality paints that caught her attention. |
| *Response:* | { "response": "No" } |
| *System Prompt:* | Does this prompt have multiple subjects or objects? If a pronoun relates to a noun in the sentence then it is considered a single subject. Respond with only a Yes or No. Only respond as a JSON object. For example, { "response": "Yes" }. |
| *Prompt:* | She carefully examined each color and texture, feeling inspired by the possibilities. |
| *Response:* | { "response": "No" } |
| | |
| *System Prompt:* | Does the given prompt imply a purchase, whether it is in the past, present, or future? Respond with a Yes or a No. Only respond as a JSON object. For example, { "response": "Yes" }. |
| *Prompt:* | She carefully examined each color and texture, feeling inspired by the possibilities. |
| *Response:* | { "response": "No" } |
| | |
| *System Prompt:* | Does this prompt have multiple subjects or objects? If a pronoun relates to a noun in the sentence then it is considered a single subject. Respond with only a Yes or No. Only respond as a JSON object. For example, { "response": "Yes" }. |
| *Prompt:* | The sales associate noticed Lily's enthusiasm and approached her with a smile. |
| *Response:* | { "response": "No" } |
| | |
| *System Prompt:* | Does the given prompt imply a purchase, whether it is in the past, present, or future? Respond with a Yes or a No. Only respond as a JSON object. For example, { "response": "Yes" }. |
| *Prompt:* | The sales associate noticed Lily's enthusiasm and approached her with a smile. |
| *Response:* | { "response": "No" } |
| | |
| *System Prompt:* | Does this prompt have multiple subjects or objects? If a pronoun relates to a noun in the sentence then it is considered a single subject. Respond with only a Yes or No. Only respond as a JSON object. For example, { "response": "Yes" }. |
| *Prompt:* | "Those are our top-of-the-line paints," he said. |
| *Response:* | { "response": "No" } |
| | |
| *System Prompt:* | Does the given prompt imply a purchase, whether it is in the past, present, or future? Respond with a Yes or a No. Only respond as a JSON object. For example, { "response": "Yes" }. |
| *Prompt:* | "Those are our top-of-the-line paints," he said. |
| *Response:* | { "response": "No" } |
| | |
| *System Prompt:* | Does this prompt have multiple subjects or objects? If a pronoun relates to a noun in the sentence then it is considered a single subject. Respond with only a Yes or No. Only respond as a JSON object. For example, { "response": "Yes" }. |
| *Prompt:* | "They're made from the finest materials and will give you incredible results." |
| *Response:* | { "response": "No" } |
| | |
| *System Prompt:* | Does the given prompt imply a purchase, whether it is in the past, present, or future? Respond with a Yes or a No. Only respond as a JSON object. For example, { "response": "Yes" }. |
| *Prompt:* | "They're made from the finest materials and will give you incredible results." |
| *Response:* | { "response": "No" } |
| | |
| *System Prompt:* | Does this prompt have multiple subjects or objects? If a pronoun relates to a noun in the sentence then it is considered a single subject. Respond with only a Yes or No. Only respond as a JSON object. For example, { "response": "Yes" }. |
| *Prompt:* | Lily couldn't resist - she bought the entire set and spent the rest of the day creating new masterpieces. |
| *Response:* | { "response": "No" } |
| | |
| *System Prompt:* | Does the given prompt imply a purchase, whether it is in the past, present, or future? Respond with a Yes or a No. Only respond as a JSON object. For example, { "response": "Yes" }. |
| *Prompt:* | Lily couldn't resist - she bought the entire set and spent the rest of the day creating new masterpieces. |
| *Response:* | { "response": "Yes" } |
| *System Prompt:* | Analyze the sentence and identify all SVO phrases. Replace any pronouns (e.g. "he" or "she" or "they" or "I" or "you") with proper nouns. Respond with ONLY a JSON object with only three fields: subject, object, and action. For example, if I say "Ben is so happy - he bought a boat and is going sailing", you should reply with { "subject": "Ben", "object": "a boat", "action": "bought" } |
| *Prompt:* | Lily couldn't resist - she bought the entire set and spent the rest of the day creating new masterpieces. |
| *Response:* | {<br>  "subject": "Lily",<br>  "object": "the entire set",<br>  "action": "bought"<br>} |

**Listing A7.** Test 7.

Start: 2:39:23 PM          End: 2:36:41 PM

| | |
|---|---|
| *System Prompt:* | Is the following input multiple independent clauses? Respond with only a Yes or No. Only respond as a JSON object. For example, "Dillan ate an apple. He thought it tasted good. " { "response": "Yes" }. |
| *Prompt:* | Alex had been using his old phone for years, but it was finally time to upgrade. As he walked into the electronics store, his eyes landed on a sleek new smartphone that caught his attention. He carefully examined each feature and function, feeling impressed by its capabilities. The sales associate noticed Alex's interest and approached him with a smile. "That's one of our latest models," he said. "It has all the latest features and is designed for speed and efficiency." Alex couldn't resist - he bought the phone on the spot and spent the rest of the day exploring its new features. |
| *Response:* | { "response": "Yes" } |
| *System Prompt:* | Split this paragraph into distinct sentences. Only respond as a JSON array For example if given the text "This is one sentence. This is another sentence", respond with [ "This is one sentence.", "This is another sentence." ]. |
| *Prompt:* | Alex had been using his old phone for years, but it was finally time to upgrade. As he walked into the electronics store, his eyes landed on a sleek new smartphone that caught his attention. He carefully examined each feature and function, feeling impressed by its capabilities. The sales associate noticed Alex's interest and approached him with a smile. \"That's one of our latest models,\" he said. \"It has all the latest features and is designed for speed and efficiency.\" Alex couldn't resist - he bought the phone on the spot and spent the rest of the day exploring its new features. |
| *Response:* | [<br>  "Alex had been using his old phone for years, but it was finally time to upgrade.",<br>  "As he walked into the electronics store, his eyes landed on a sleek new smartphone that caught his attention.",<br>  "He carefully examined each feature and function, feeling impressed by its capabilities.",<br>  "The sales associate noticed Alex's interest and approached him with a smile.",<br>  "\"That's one of our latest models,\" he said.",<br>  "\"It has all the latest features and is designed for speed and efficiency.\"",<br>  "Alex couldn't resist - he bought the phone on the spot and spent the rest of the day exploring its new features."<br>] |
| *System Prompt:* | Does this prompt have multiple subjects or objects? If a pronoun relates to a noun in the sentence then it is considered a single subject. Respond with only a Yes or No. Only respond as a JSON object. For example, { "response": "Yes" }. |
| *Prompt:* | Alex had been using his old phone for years, but it was finally time to upgrade. |
| *Response:* | { "response": "No" } |
| *System Prompt:* | Does the given prompt imply a purchase, whether it is in the past, present, or future? Respond with a Yes or a No. Only respond as a JSON object. For example, { "response": "Yes" }. |
| *Prompt:* | Alex had been using his old phone for years, but it was finally time to upgrade. |
| *Response:* | { "response": "No" } |
| | |
| *System Prompt:* | Does this prompt have multiple subjects or objects? If a pronoun relates to a noun in the sentence then it is considered a single subject. Respond with only a Yes or No. Only respond as a JSON object. For example, { "response": "Yes" }. |
| *Prompt:* | As he walked into the electronics store, his eyes landed on a sleek new smartphone that caught his attention. |
| *Response:* | { "response": "No" } |
| | |
| *System Prompt:* | Does the given prompt imply a purchase, whether it is in the past, present, or future? Respond with a Yes or a No. Only respond as a JSON object. For example, { "response": "Yes" }. |
| *Prompt:* | As he walked into the electronics store, his eyes landed on a sleek new smartphone that caught his attention. |
| *Response:* | { "response": "No" } |
| | |
| *System Prompt:* | Does this prompt have multiple subjects or objects? If a pronoun relates to a noun in the sentence then it is considered a single subject. Respond with only a Yes or No. Only respond as a JSON object. For example, { "response": "Yes" }. |
| *Prompt:* | He carefully examined each feature and function, feeling impressed by its capabilities. |
| *Response:* | { "response": "No" } |
| | |
| *System Prompt:* | Does the given prompt imply a purchase, whether it is in the past, present, or future? Respond with a Yes or a No. Only respond as a JSON object. For example, { "response": "Yes" }. |
| *Prompt:* | He carefully examined each feature and function, feeling impressed by its capabilities. |
| *Response:* | { "response": "No" } |
| | |
| *System Prompt:* | Does this prompt have multiple subjects or objects? If a pronoun relates to a noun in the sentence then it is considered a single subject. Respond with only a Yes or No. Only respond as a JSON object. For example, { "response": "Yes" }. |
| *Prompt:* | The sales associate noticed Alex's interest and approached him with a smile. |
| *Response:* | { "response": "No" } |
| | |
| *System Prompt:* | Does the given prompt imply a purchase, whether it is in the past, present, or future? Respond with a Yes or a No. Only respond as a JSON object. For example, { "response": "Yes" }. |
| *Prompt:* | The sales associate noticed Alex's interest and approached him with a smile. |
| *Response:* | { "response": "No" } |
| | |
| *System Prompt:* | Does this prompt have multiple subjects or objects? If a pronoun relates to a noun in the sentence then it is considered a single subject. Respond with only a Yes or No. Only respond as a JSON object. For example, { "response": "Yes" }. |
| *Prompt:* | "That's one of our latest models," he said. |
| *Response:* | { "response": "No" } |
| | |
| *System Prompt:* | Does the given prompt imply a purchase, whether it is in the past, present, or future? Respond with a Yes or a No. Only respond as a JSON object. For example, { "response": "Yes" }. |
| *Prompt:* | "That's one of our latest models," he said. |
| *Response:* | { "response": "No" } |
| | |
| *System Prompt:* | Does this prompt have multiple subjects or objects? If a pronoun relates to a noun in the sentence then it is considered a single subject. Respond with only a Yes or No. Only respond as a JSON object. For example, { "response": "Yes" }. |
| *Prompt:* | "It has all the latest features and is designed for speed and efficiency." |
| *Response:* | { "response": "No" } |
| *System Prompt:* | Does the given prompt imply a purchase, whether it is in the past, present, or future? Respond with a Yes or a No. Only respond as a JSON object. For example, { "response": "Yes" }. |
| *Prompt:* | "It has all the latest features and is designed for speed and efficiency." |
| *Response:* | { "response": "No" } |
| *System Prompt:* | Does this prompt have multiple subjects or objects? If a pronoun relates to a noun in the sentence then it is considered a single subject. Respond with only a Yes or No. Only respond as a JSON object. For example, { "response": "Yes" }. |
| *Prompt:* | Alex couldn't resist - he bought the phone on the spot and spent the rest of the day exploring its new features. |
| *Response:* | { "response": "No" } |
| *System Prompt:* | Does the given prompt imply a purchase, whether it is in the past, present, or future? Respond with a Yes or a No. Only respond as a JSON object. For example, { "response": "Yes" }. |
| *Prompt:* | Alex couldn't resist - he bought the phone on the spot and spent the rest of the day exploring its new features. |
| *Response:* | { "response": "Yes" } |
| *System Prompt:* | Analyze the sentence and identify all SVO phrases. Replace any pronouns (e.g. "he" or "she" or "they" or "I" or "you") with proper nouns. Respond with ONLY a JSON object with only three fields: subject, object, and action. For example, if I say "Ben is so happy - he bought a boat and is going sailing", you should reply with { "subject": "Ben", "object": "a boat", "action": "bought" } |
| *Prompt:* | Alex couldn't resist - he bought the phone on the spot and spent the rest of the day exploring its new features. |
| *Response:* | {<br>  "subject": "Alex",<br>  "object": "the phone",<br>  "action": "bought"<br>} |

**Listing A8.** Test 8.
Start: 2:40:30 PM          End: Didn't Complete - Error

| | |
|---|---|
| *System Prompt:* | Is the following input multiple independent clauses? Respond with only a Yes or No. Only respond as a JSON object. For example, "Dillan ate an apple. He thought it tasted good. " { "response": "Yes" }. |
| *Prompt:* | Sarah had always loved gardening, but she was tired of using old, rusty tools that didn't get the job done. As she walked into the garden center, her eyes landed on a set of high-quality gardening tools that caught her attention. She carefully examined each tool and felt impressed by its durability and quality. The sales associate noticed Sarah's enthusiasm and approached her with a smile. "Those are our top-of-the-line tools," he said. "They're made from the finest materials and will make your gardening tasks easier." Sarah couldn't resist - she bought the entire set and spent the rest of the day tending to her garden. |
| *Response:* | { "response": "Yes" } |
| *System Prompt:* | Split this paragraph into distinct sentences. Only respond as a JSON array For example if given the text "This is one sentence. This is another sentence", respond with [ "This is one sentence.", "This is another sentence." ]. |
| *Prompt:* | Sarah had always loved gardening, but she was tired of using old, rusty tools that didn't get the job done. As she walked into the garden center, her eyes landed on a set of high-quality gardening tools that caught her attention. She carefully examined each tool and felt impressed by its durability and quality. The sales associate noticed Sarah's enthusiasm and approached her with a smile. \"Those are our top-of-the-line tools,\" he said. \"They're made from the finest materials and will make your gardening tasks easier.\" Sarah couldn't resist - she bought the entire set and spent |
| | the rest of the day tending to her garden. |
| *Response:* | [<br>   "Sarah had always loved gardening, but she was tired of using old, rusty tools that didn't get the job done.",<br>   "As she walked into the garden center, her eyes landed on a set of high-quality gardening tools that caught her attention.",<br>   "She carefully examined each tool and felt impressed by its durability and quality.",<br>   "The sales associate noticed Sarah's enthusiasm and approached her with a smile.",<br>   "\"Those are our top-of-the-line tools,\" he said.",<br>   "\"They're made from the finest materials and will make your gardening tasks easier.\"",<br>   "Sarah couldn't resist - she bought the entire set and spent the rest of the day tending to her garden."<br>] |
| *System Prompt:* | Does this prompt have multiple subjects or objects? If a pronoun relates to a noun in the sentence then it is considered a single subject. Respond with only a Yes or No. Only respond as a JSON object. For example, { "response": "Yes" }. |
| *Prompt:* | Sarah had always loved gardening, but she was tired of using old, rusty tools that didn't get the job done. |
| *Response:* | { "response": "No" } |
| *System Prompt:* | Does the given prompt imply a purchase, whether it is in the past, present, or future? Respond with a Yes or a No. Only respond as a JSON object. For example, { "response": "Yes" }. |
| *Prompt:* | Sarah had always loved gardening, but she was tired of using old, rusty tools that didn't get the job done. |
| *Response:* | { "response": "No" } |
| *System Prompt:* | Does this prompt have multiple subjects or objects? If a pronoun relates to a noun in the sentence then it is considered a single subject. Respond with only a Yes or No. Only respond as a JSON object. For example, { "response": "Yes" }. |
| *Prompt:* | As she walked into the garden center, her eyes landed on a set of high-quality gardening tools that caught her attention. |
| *Response:* | { "response": "No" } |
| *System Prompt:* | Does the given prompt imply a purchase, whether it is in the past, present, or future? Respond with a Yes or a No. Only respond as a JSON object. For example, { "response": "Yes" }. |
| *Prompt:* | As she walked into the garden center, her eyes landed on a set of high-quality gardening tools that caught her attention. |
| *Response:* | { "response": "No" } |
| *System Prompt:* | Does this prompt have multiple subjects or objects? If a pronoun relates to a noun in the sentence then it is considered a single subject. Respond with only a Yes or No. Only respond as a JSON object. For example, { "response": "Yes" }. |
| *Prompt:* | She carefully examined each tool and felt impressed by its durability and quality. |
| *Response:* | { "response": "No" } |
| *System Prompt:* | Does the given prompt imply a purchase, whether it is in the past, present, or future? Respond with a Yes or a No. Only respond as a JSON object. For example, { "response": "Yes" }. |
| *Prompt:* | She carefully examined each tool and felt impressed by its durability and quality. |
| *Response:* | { "response": "No" } |
| *System Prompt:* | Does this prompt have multiple subjects or objects? If a pronoun relates to a noun in the sentence then it is considered a single subject. Respond with only a Yes or No. Only respond as a JSON object. For example, { "response": "Yes" }. |
| *Prompt:* | The sales associate noticed Sarah's enthusiasm and approached her with a smile. |
| *Response:* | { "response": "No" } |
| *System Prompt:* | Does the given prompt imply a purchase, whether it is in the past, present, or future? Respond with a Yes or a No. Only respond as a JSON object. For example, { "response": "Yes" }. |
| *Prompt:* | The sales associate noticed Sarah's enthusiasm and approached her with a smile. |
| *Response:* | { "response": "No" } |
| | |
| *System Prompt:* | Does this prompt have multiple subjects or objects? If a pronoun relates to a noun in the sentence then it is considered a single subject. Respond with only a Yes or No. Only respond as a JSON object. For example, { "response": "Yes" }. |
| *Prompt:* | "Those are our top-of-the-line tools," he said. |
| *Response:* | { "response": "No" } |
| | |
| *System Prompt:* | Does the given prompt imply a purchase, whether it is in the past, present, or future? Respond with a Yes or a No. Only respond as a JSON object. For example, { "response": "Yes" }. |
| *Prompt:* | "Those are our top-of-the-line tools," he said. |
| *Response:* | { "response": "No" } |
| | |
| *System Prompt:* | Does this prompt have multiple subjects or objects? If a pronoun relates to a noun in the sentence then it is considered a single subject. Respond with only a Yes or No. Only respond as a JSON object. For example, { "response": "Yes" }. |
| *Prompt:* | "They're made from the finest materials and will make your gardening tasks easier." |
| *Response:* | { "response": "No" } |
| | |
| *System Prompt:* | Does the given prompt imply a purchase, whether it is in the past, present, or future? Respond with a Yes or a No. Only respond as a JSON object. For example, { "response": "Yes" }. |
| *Prompt:* | "They're made from the finest materials and will make your gardening tasks easier." |
| *Response:* | { "response": "No" } |
| | |
| *System Prompt:* | Does this prompt have multiple subjects or objects? If a pronoun relates to a noun in the sentence then it is considered a single subject. Respond with only a Yes or No. Only respond as a JSON object. For example, { "response": "Yes" }. |
| *Prompt:* | Sarah couldn't resist - she bought the entire set and spent the rest of the day tending to her garden. |
| *Response:* | { "response": "No" } |
| | |
| *System Prompt:* | Does the given prompt imply a purchase, whether it is in the past, present, or future? Respond with a Yes or a No. Only respond as a JSON object. For example, { "response": "Yes" }. |
| *Prompt:* | Sarah couldn't resist - she bought the entire set and spent the rest of the day tending to her garden. |
| *Response:* | { "response": "Yes" } |
| | |
| *System Prompt:* | Analyze the sentence and identify all SVO phrases. Replace any pronouns (e.g. "he" or "she" or "they" or "I" or "you") with proper nouns. Respond with ONLY a JSON object with only three fields: subject, object, and action. For example, if I say "Ben is so happy - he bought a boat and is going sailing", you should reply with { "subject": "Ben", "object": "a boat", "action": "bought" } |
| *Prompt:* | Sarah couldn't resist - she bought the entire set and spent the rest of the day tending to her garden. |
| *Response:* | {<br>  "subject": "Sarah",<br>  "object": "the entire set",<br>  "action": "bought"<br>} |

**Listing A9.** Test 9.
Start: 2:41:30 PM      End: 2:41:52 PM

| | |
|---|---|
| *System Prompt:* | Is the following input multiple independent clauses? Respond with only a Yes or No. Only respond as a JSON object. For example, "Dillan ate an apple. He thought it tasted good. " { "response": "Yes" }. |
| *Prompt:* | Emily was a student, and she needed a reliable laptop for her studies. As she walked into the electronics store, her eyes landed on a sleek new laptop that caught her attention. She carefully examined each feature and function, feeling impressed by its capabilities. The sales associate noticed Emily's interest and approached her with a smile. "That's one of our latest models," he said. "It has all the latest features and is designed for speed and efficiency." Emily couldn't resist - she bought the laptop on the spot and spent the rest of the day setting it up and exploring its new features. |
| *Response:* | { "response": "Yes" } |
| *System Prompt:* | Split this paragraph into distinct sentences. Only respond as a JSON array For example if given the text "This is one sentence. This is another sentence", respond with [ "This is one sentence.", "This is another sentence." ]. |
| *Prompt:* | Emily was a student, and she needed a reliable laptop for her studies. As she walked into the electronics store, her eyes landed on a sleek new laptop that caught her attention. She carefully examined each feature and function, feeling impressed by its capabilities. The sales associate noticed Emily's interest and approached her with a smile. \"That's one of our latest models,\" he said. \"It has all the latest features and is designed for speed and efficiency.\" Emily couldn't resist - she bought the laptop on the spot and spent the rest of the day setting it up and exploring its new features. |
| *Response:* | [<br>  "Emily was a student, and she needed a reliable laptop for her studies.",<br>  "As she walked into the electronics store, her eyes landed on a sleek new laptop that caught her attention.",<br>  "She carefully examined each feature and function, feeling impressed by its capabilities.",<br>  "The sales associate noticed Emily's interest and approached her with a smile.",<br>  "\"That's one of our latest models,\" he said.",<br>  "\"It has all the latest features and is designed for speed and efficiency.\"",<br>  "Emily couldn't resist - she bought the laptop on the spot and spent the rest of the day setting it up and exploring its new features."<br>] |
| *System Prompt:* | Does this prompt have multiple subjects or objects? If a pronoun relates to a noun in the sentence then it is considered a single subject. Respond with only a Yes or No. Only respond as a JSON object. For example, { "response": "Yes" }. |
| *Prompt:* | Emily was a student, and she needed a reliable laptop for her studies. |
| *Response:* | { "response": "No" } |
| *System Prompt:* | Does the given prompt imply a purchase, whether it is in the past, present, or future? Respond with a Yes or a No. Only respond as a JSON object. For example, { "response": "Yes" }. |
| *Prompt:* | Emily was a student, and she needed a reliable laptop for her studies. |
| *Response:* | { "response": "No" } |
| *System Prompt:* | Does this prompt have multiple subjects or objects? If a pronoun relates to a noun in the sentence then it is considered a single subject. Respond with only a Yes or No. Only respond as a JSON object. For example, { "response": "Yes" }. |
| *Prompt:* | As she walked into the electronics store, her eyes landed on a sleek new laptop that caught her attention. |
| *Response:* | { "response": "No" } |
| *System Prompt:* | Does the given prompt imply a purchase, whether it is in the past, present, or future? Respond with a Yes or a No. Only respond as a JSON object. For example, { "response": "Yes" }. |
| *Prompt:* | As she walked into the electronics store, her eyes landed on a sleek new laptop that caught her attention. |
| *Response:* | { "response": "No" } |
| *System Prompt:* | Does this prompt have multiple subjects or objects? If a pronoun relates to a noun in the sentence |
| | then it is considered a single subject. Respond with only a Yes or No. Only respond as a JSON object. For example, { "response": "Yes" }. |
| *Prompt:* | She carefully examined each feature and function, feeling impressed by its capabilities. |
| *Response:* | { "response": "No" } |
| | |
| *System Prompt:* | Does the given prompt imply a purchase, whether it is in the past, present, or future? Respond with a Yes or a No. Only respond as a JSON object. For example, { "response": "Yes" }. |
| *Prompt:* | She carefully examined each feature and function, feeling impressed by its capabilities. |
| *Response:* | { "response": "No" } |
| | |
| *System Prompt:* | Does this prompt have multiple subjects or objects? If a pronoun relates to a noun in the sentence then it is considered a single subject. Respond with only a Yes or No. Only respond as a JSON object. For example, { "response": "Yes" }. |
| *Prompt:* | The sales associate noticed Emily's interest and approached her with a smile. |
| *Response:* | { "response": "No" } |
| | |
| *System Prompt:* | Does the given prompt imply a purchase, whether it is in the past, present, or future? Respond with a Yes or a No. Only respond as a JSON object. For example, { "response": "Yes" }. |
| *Prompt:* | The sales associate noticed Emily's interest and approached her with a smile. |
| *Response:* | { "response": "No" } |
| | |
| *System Prompt:* | Does this prompt have multiple subjects or objects? If a pronoun relates to a noun in the sentence then it is considered a single subject. Respond with only a Yes or No. Only respond as a JSON object. For example, { "response": "Yes" }. |
| *Prompt:* | "That's one of our latest models," he said. |
| *Response:* | { "response": "No" } |
| | |
| *System Prompt:* | Does the given prompt imply a purchase, whether it is in the past, present, or future? Respond with a Yes or a No. Only respond as a JSON object. For example, { "response": "Yes" }. |
| *Prompt:* | "That's one of our latest models," he said. |
| *Response:* | { "response": "No" } |
| | |
| *System Prompt:* | Does this prompt have multiple subjects or objects? If a pronoun relates to a noun in the sentence then it is considered a single subject. Respond with only a Yes or No. Only respond as a JSON object. For example, { "response": "Yes" }. |
| *Prompt:* | "It has all the latest features and is designed for speed and efficiency." |
| *Response:* | { "response": "No" } |
| | |
| *System Prompt:* | Does the given prompt imply a purchase, whether it is in the past, present, or future? Respond with a Yes or a No. Only respond as a JSON object. For example, { "response": "Yes" }. |
| *Prompt:* | "It has all the latest features and is designed for speed and efficiency." |
| *Response:* | { "response": "No" } |
| | |
| *System Prompt:* | Does this prompt have multiple subjects or objects? If a pronoun relates to a noun in the sentence then it is considered a single subject. Respond with only a Yes or No. Only respond as a JSON object. For example, { "response": "Yes" }. |
| *Prompt:* | Emily couldn't resist - she bought the laptop on the spot and spent the rest of the day setting it up and exploring its new features. |
| *Response:* | { "response": "No" } |
| | |
| *System Prompt:* | Does the given prompt imply a purchase, whether it is in the past, present, or future? Respond with a Yes or a No. Only respond as a JSON object. For example, { "response": "Yes" }. |
| *Prompt:* | Emily couldn't resist - she bought the laptop on the spot and spent the rest of the day setting it up and exploring its new features. |
| *Response:* | { "response": "Yes" } |

*System Prompt:* Analyze the sentence and identify all SVO phrases. Replace any pronouns (e.g. "he" or "she" or "they" or "I" or "you") with proper nouns. Respond with ONLY a JSON object with only three fields: subject, object, and action. For example, if I say "Ben is so happy - he bought a boat and is going sailing", you should reply with { "subject": "Ben", "object": "a boat", "action": "bought" }

*Prompt:* Emily couldn't resist - she bought the laptop on the spot and spent the rest of the day setting it up and exploring its new features.

*Response:*
```
{
  "subject": "Emily",
  "object": "the laptop",
  "action": "bought"
}
```

**Listing A10.** Test 10.

Start: 2:42:28 PM          End: Didn't Complete - Error

*System Prompt:* Is the following input multiple independent clauses? Respond with only a Yes or No. Only respond as a JSON object. For example, "Dillan ate an apple. He thought it tasted good. " { "response": "Yes" }.

*Prompt:* Rachel was celebrating her anniversary with her husband, and she wanted to find the perfect gift. As she walked into the jewelry store, her eyes landed on a beautiful set of fine jewelry that caught her attention. She carefully examined each piece and felt impressed by its quality and craftsmanship. The sales associate noticed Rachel's enthusiasm and approached her with a smile. "That's one of our finest collections," he said. "It's made from high-quality materials and will last for generations." Rachel couldn't resist - she bought the entire set and spent the rest of the day planning the perfect anniversary celebration.

*Response:* { "response": "Yes" }

*System Prompt:* Split this paragraph into distinct sentences. Only respond as a JSON array For example if given the text "This is one sentence. This is another sentence", respond with [ "This is one sentence.", "This is another sentence." ].

*Prompt:* Rachel was celebrating her anniversary with her husband, and she wanted to find the perfect gift. As she walked into the jewelry store, her eyes landed on a beautiful set of fine jewelry that caught her attention. She carefully examined each piece and felt impressed by its quality and craftsmanship. The sales associate noticed Rachel's enthusiasm and approached her with a smile. \"That's one of our finest collections,\" he said. \"It's made from high-quality materials and will last for generations.\" Rachel couldn't resist - she bought the entire set and spent the rest of the day planning the perfect anniversary celebration.

*Response:*
```
[
   "Rachel was celebrating her anniversary with her husband, and she wanted to find the perfect gift.",
   "As she walked into the jewelry store, her eyes landed on a beautiful set of fine jewelry that caught her attention.",
   "She carefully examined each piece and felt impressed by its quality and craftsmanship.",
   "The sales associate noticed Rachel's enthusiasm and approached her with a smile.",
   "\"That's one of our finest collections,\" he said.",
   "\"It's made from high-quality materials and will last for generations.\"",
   "Rachel couldn't resist - she bought the entire set and spent the rest of the day planning the perfect anniversary celebration."
]
```

*System Prompt:* Does this prompt have multiple subjects or objects? If a pronoun relates to a noun in the sentence then it is considered a single subject. Respond with only a Yes or No. Only respond as a JSON object. For example, { "response": "Yes" }.

*Prompt:* Rachel was celebrating her anniversary with her husband, and she wanted to find the perfect gift.

*Response:* { "response": "No" }

| | |
|---|---|
| *System Prompt:* | Does the given prompt imply a purchase, whether it is in the past, present, or future? Respond with a Yes or a No. Only respond as a JSON object. For example, { "response": "Yes" }. |
| *Prompt:* | Rachel was celebrating her anniversary with her husband, and she wanted to find the perfect gift. |
| *Response:* | { "response": "No" } |
| | |
| *System Prompt:* | Does this prompt have multiple subjects or objects? If a pronoun relates to a noun in the sentence then it is considered a single subject. Respond with only a Yes or No. Only respond as a JSON object. For example, { "response": "Yes" }. |
| *Prompt:* | As she walked into the jewelry store, her eyes landed on a beautiful set of fine jewelry that caught her attention. |
| *Response:* | { "response": "No" } |
| | |
| *System Prompt:* | Does the given prompt imply a purchase, whether it is in the past, present, or future? Respond with a Yes or a No. Only respond as a JSON object. For example, { "response": "Yes" }. |
| *Prompt:* | As she walked into the jewelry store, her eyes landed on a beautiful set of fine jewelry that caught her attention. |
| *Response:* | { "response": "No" } |
| | |
| *System Prompt:* | Does this prompt have multiple subjects or objects? If a pronoun relates to a noun in the sentence then it is considered a single subject. Respond with only a Yes or No. Only respond as a JSON object. For example, { "response": "Yes" }. |
| *Prompt:* | She carefully examined each piece and felt impressed by its quality and craftsmanship. |
| *Response:* | { "response": "No" } |
| | |
| *System Prompt:* | Does the given prompt imply a purchase, whether it is in the past, present, or future? Respond with a Yes or a No. Only respond as a JSON object. For example, { "response": "Yes" }. |
| *Prompt:* | She carefully examined each piece and felt impressed by its quality and craftsmanship. |
| *Response:* | { "response": "No" } |
| | |
| *System Prompt:* | Does this prompt have multiple subjects or objects? If a pronoun relates to a noun in the sentence then it is considered a single subject. Respond with only a Yes or No. Only respond as a JSON object. For example, { "response": "Yes" }. |
| *Prompt:* | The sales associate noticed Rachel's enthusiasm and approached her with a smile. |
| *Response:* | { "response": "No" } |
| | |
| *System Prompt:* | Does the given prompt imply a purchase, whether it is in the past, present, or future? Respond with a Yes or a No. Only respond as a JSON object. For example, { "response": "Yes" }. |
| *Prompt:* | The sales associate noticed Rachel's enthusiasm and approached her with a smile. |
| *Response:* | { "response": "No" } |
| | |
| *System Prompt:* | Does this prompt have multiple subjects or objects? If a pronoun relates to a noun in the sentence then it is considered a single subject. Respond with only a Yes or No. Only respond as a JSON object. For example, { "response": "Yes" }. |
| *Prompt:* | "That's one of our finest collections," he said. |
| *Response:* | { "response": "No" } |
| | |
| *System Prompt:* | Does the given prompt imply a purchase, whether it is in the past, present, or future? Respond with a Yes or a No. Only respond as a JSON object. For example, { "response": "Yes" }. |
| *Prompt:* | "That's one of our finest collections," he said. |
| *Response:* | { "response": "No" } |
| | |
| *System Prompt:* | Does this prompt have multiple subjects or objects? If a pronoun relates to a noun in the sentence then it is considered a single subject. Respond with only a Yes or No. Only respond as a JSON object. For example, { "response": "Yes" }. |
| *Prompt:* | "It's made from high-quality materials and will last for generations." |
| *Response:* | { "response": "No" } |
| *System Prompt:* | Does the given prompt imply a purchase, whether it is in the past, present, or future? Respond with a Yes or a No. Only respond as a JSON object. For example, { "response": "Yes" }. |
| *Prompt:* | "It's made from high-quality materials and will last for generations." |
| *Response:* | { "response": "No" } |
| | |
| *System Prompt:* | Does this prompt have multiple subjects or objects? If a pronoun relates to a noun in the sentence then it is considered a single subject. Respond with only a Yes or No. Only respond as a JSON object. For example, { "response": "Yes" }. |
| *Prompt:* | Rachel couldn't resist - she bought the entire set and spent the rest of the day planning the perfect anniversary celebration. |
| *Response:* | { "response": "No" } |
| | |
| *System Prompt:* | Does the given prompt imply a purchase, whether it is in the past, present, or future? Respond with a Yes or a No. Only respond as a JSON object. For example, { "response": "Yes" }. |
| *Prompt:* | Rachel couldn't resist - she bought the entire set and spent the rest of the day planning the perfect anniversary celebration. |
| *Response:* | { "response": "Yes" } |
| | |
| *System Prompt:* | Analyze the sentence and identify all SVO phrases. Replace any pronouns (e.g. "he" or "she" or "they" or "I" or "you") with proper nouns. Respond with ONLY a JSON object with only three fields: subject, object, and action. For example, if I say "Ben is so happy - he bought a boat and is going sailing", you should reply with { "subject": "Ben", "object": "a boat", "action": "bought" } |
| *Prompt:* | Rachel couldn't resist - she bought the entire set and spent the rest of the day planning the perfect anniversary celebration. |
| *Response:* | {<br>  "subject": "Rachel",<br>  "object": "the entire set",<br>  "action": "bought"<br>} |

**Listing A11.** Test 11.
Start: 2:43:26 PM         End: 2:43:50 PM

| | |
|---|---|
| *System Prompt:* | Is the following input multiple independent clauses? Respond with only a Yes or No. Only respond as a JSON object. For example, "Dillan ate an apple. He thought it tasted good. " { "response": "Yes" }. |
| *Prompt:* | Jack had always loved photography, but he was tired of using an old camera that didn't get the job done. As he walked into the electronics store, his eyes landed on a sleek new camera that caught his attention. He carefully examined each feature and function, feeling impressed by its capabilities. The sales associate noticed Jack's enthusiasm and approached him with a smile. "That's one of our top-of-the-line cameras," he said. "It has all the latest features and is designed for speed and efficiency." Jack couldn't resist - he bought the camera on the spot and spent the rest of the day exploring its new features and taking stunning photos. |
| *Response:* | {<br>  "response": "Yes"<br>} |
| | |
| *System Prompt:* | Split this paragraph into distinct sentences. Only respond as a JSON array For example if given the text "This is one sentence. This is another sentence", respond with [ "This is one sentence.", "This is another sentence." ]. |
| *Prompt:* | Jack had always loved photography, but he was tired of using an old camera that didn't get the job done. As he walked into the electronics store, his eyes landed on a sleek new camera that caught his attention. He carefully examined each feature and function, feeling impressed by its capabilities. The sales associate noticed Jack's enthusiasm and approached him with a smile. \"That's one of our top-of-the-line cameras,\" he said. \"It has all the latest features and is designed |
| | for speed and efficiency.\" Jack couldn't resist - he bought the camera on the spot and spent the rest of the day exploring its new features and taking stunning photos. |
| *Response:* | [<br>  "Jack had always loved photography, but he was tired of using an old camera that didn't get the job done.",<br>  "As he walked into the electronics store, his eyes landed on a sleek new camera that caught his attention.",<br>  "He carefully examined each feature and function, feeling impressed by its capabilities.",<br>  "The sales associate noticed Jack's enthusiasm and approached him with a smile.",<br>  "\"That's one of our top-of-the-line cameras,\" he said.",<br>  "\"It has all the latest features and is designed for speed and efficiency.\"",<br>  "Jack couldn't resist - he bought the camera on the spot and spent the rest of the day exploring its new features and taking stunning photos."<br>] |
| *System Prompt:* | Does this prompt have multiple subjects or objects? If a pronoun relates to a noun in the sentence then it is considered a single subject. Respond with only a Yes or No. Only respond as a JSON object. For example, { "response": "Yes" }. |
| *Prompt:* | Jack had always loved photography, but he was tired of using an old camera that didn't get the job done. |
| *Response:* | { "response": "No" } |
| *System Prompt:* | Does the given prompt imply a purchase, whether it is in the past, present, or future? Respond with a Yes or a No. Only respond as a JSON object. For example, { "response": "Yes" }. |
| *Prompt:* | Jack had always loved photography, but he was tired of using an old camera that didn't get the job done. |
| *Response:* | { "response": "No" } |
| *System Prompt:* | Does this prompt have multiple subjects or objects? If a pronoun relates to a noun in the sentence then it is considered a single subject. Respond with only a Yes or No. Only respond as a JSON object. For example, { "response": "Yes" }. |
| *Prompt:* | As he walked into the electronics store, his eyes landed on a sleek new camera that caught his attention. |
| *Response:* | { "response": "No" } |
| *System Prompt:* | Does the given prompt imply a purchase, whether it is in the past, present, or future? Respond with a Yes or a No. Only respond as a JSON object. For example, { "response": "Yes" }. |
| *Prompt:* | As he walked into the electronics store, his eyes landed on a sleek new camera that caught his attention. |
| *Response:* | { "response": "No" } |
| *System Prompt:* | Does this prompt have multiple subjects or objects? If a pronoun relates to a noun in the sentence then it is considered a single subject. Respond with only a Yes or No. Only respond as a JSON object. For example, { "response": "Yes" }. |
| *Prompt:* | He carefully examined each feature and function, feeling impressed by its capabilities. |
| *Response:* | { "response": "No" } |
| *System Prompt:* | Does the given prompt imply a purchase, whether it is in the past, present, or future? Respond with a Yes or a No. Only respond as a JSON object. For example, { "response": "Yes" }. |
| *Prompt:* | He carefully examined each feature and function, feeling impressed by its capabilities. |
| *Response:* | { "response": "No" } |
| *System Prompt:* | Does this prompt have multiple subjects or objects? If a pronoun relates to a noun in the sentence then it is considered a single subject. Respond with only a Yes or No. Only respond as a JSON object. For example, { "response": "Yes" }. |
| *Prompt:* | The sales associate noticed Jack's enthusiasm and approached him with a smile. |
| *Response:* | { "response": "No" } |
| *System Prompt:* | Does the given prompt imply a purchase, whether it is in the past, present, or future? Respond with a Yes or a No. Only respond as a JSON object. For example, { "response": "Yes" }. |
| *Prompt:* | The sales associate noticed Jack's enthusiasm and approached him with a smile. |
| *Response:* | { "response": "No" } |
| *System Prompt:* | Does this prompt have multiple subjects or objects? If a pronoun relates to a noun in the sentence then it is considered a single subject. Respond with only a Yes or No. Only respond as a JSON object. For example, { "response": "Yes" }. |
| *Prompt:* | "That's one of our top-of-the-line cameras," he said. |
| *Response:* | { "response": "No" } |
| *System Prompt:* | Does the given prompt imply a purchase, whether it is in the past, present, or future? Respond with a Yes or a No. Only respond as a JSON object. For example, { "response": "Yes" }. |
| *Prompt:* | "That's one of our top-of-the-line cameras," he said. |
| *Response:* | { "response": "No" } |
| *System Prompt:* | Does this prompt have multiple subjects or objects? If a pronoun relates to a noun in the sentence then it is considered a single subject. Respond with only a Yes or No. Only respond as a JSON object. For example, { "response": "Yes" }. |
| *Prompt:* | "It has all the latest features and is designed for speed and efficiency." |
| *Response:* | { "response": "No" } |
| *System Prompt:* | Does the given prompt imply a purchase, whether it is in the past, present, or future? Respond with a Yes or a No. Only respond as a JSON object. For example, { "response": "Yes" }. |
| *Prompt:* | "It has all the latest features and is designed for speed and efficiency." |
| *Response:* | { "response": "No" } |
| *System Prompt:* | Does this prompt have multiple subjects or objects? If a pronoun relates to a noun in the sentence then it is considered a single subject. Respond with only a Yes or No. Only respond as a JSON object. For example, { "response": "Yes" }. |
| *Prompt:* | Jack couldn't resist - he bought the camera on the spot and spent the rest of the day exploring its new features and taking stunning photos. |
| *Response:* | { "response": "No" } |
| *System Prompt:* | Does the given prompt imply a purchase, whether it is in the past, present, or future? Respond with a Yes or a No. Only respond as a JSON object. For example, { "response": "Yes" }. |
| *Prompt:* | Jack couldn't resist - he bought the camera on the spot and spent the rest of the day exploring its new features and taking stunning photos. |
| *Response:* | { "response": "Yes" } |
| *System Prompt:* | Analyze the sentence and identify all SVO phrases. Replace any pronouns (e.g. "he" or "she" or "they" or "I" or "you") with proper nouns. Respond with ONLY a JSON object with only three fields: subject, object, and action. For example, if I say "Ben is so happy - he bought a boat and is going sailing", you should reply with { "subject": "Ben", "object": "a boat", "action": "bought" } |
| *Prompt:* | Jack couldn't resist - he bought the camera on the spot and spent the rest of the day exploring its new features and taking stunning photos. |
| *Response:* | {<br>  "subject": "Jack",<br>  "object": "the camera",<br>  "action": "bought"<br>} |

**Listing A12.** Test 12.
Start: 2:44:38 PM        End: 2:45:09 PM

| | |
|---|---|
| *System Prompt:* | Is the following input multiple independent clauses? Respond with only a Yes or No. Only respond as a JSON object. For example, "Dillan ate an apple. He thought it tasted good. " { "response": "Yes" }. |
| *Prompt:* | As I soared above the Serengeti National Park in a hot air balloon, I felt a sense of wonder and awe that I had never experienced before. The sun was rising over the savannah, casting a golden glow over the grasslands and acacia trees. Below me, the majestic animals of Africa roamed freely - giraffes, zebras, and lions all seemed to be going about their day as if I wasn't there at all. Our pilot expertly navigated the balloon through the wispy clouds, offering breathtaking views of the landscape below. As we floated above a herd of grazing wildebeest, I felt a sense of peace wash over me - this was truly a once-in-a-lifetime experience. |
| *Response:* | { "response": "Yes" } |
| *System Prompt:* | Split this paragraph into distinct sentences. Only respond as a JSON array For example if given the text "This is one sentence. This is another sentence", respond with [ "This is one sentence.", "This is another sentence." ]. |
| *Prompt:* | As I soared above the Serengeti National Park in a hot air balloon, I felt a sense of wonder and awe that I had never experienced before. The sun was rising over the savannah, casting a golden glow over the grasslands and acacia trees. Below me, the majestic animals of Africa roamed freely - giraffes, zebras, and lions all seemed to be going about their day as if I wasn't there at all. Our pilot expertly navigated the balloon through the wispy clouds, offering breathtaking views of the landscape below. As we floated above a herd of grazing wildebeest, I felt a sense of peace wash over me - this was truly a once-in-a-lifetime experience. |
| *Response:* | [<br>  "As I soared above the Serengeti National Park in a hot air balloon, I felt a sense of wonder and awe that I had never experienced before.",<br>  "The sun was rising over the savannah, casting a golden glow over the grasslands and acacia trees.",<br>  "Below me, the majestic animals of Africa roamed freely - giraffes, zebras, and lions all seemed to be going about their day as if I wasn't there at all.",<br>  "Our pilot expertly navigated the balloon through the wispy clouds, offering breathtaking views of the landscape below.",<br>  "As we floated above a herd of grazing wildebeest, I felt a sense of peace wash over me - this was truly a once-in-a-lifetime experience."<br>] |
| *System Prompt:* | Does this prompt have multiple subjects or objects? If a pronoun relates to a noun in the sentence then it is considered a single subject. Respond with only a Yes or No. Only respond as a JSON object. For example, { "response": "Yes" }. |
| *Prompt:* | As I soared above the Serengeti National Park in a hot air balloon, I felt a sense of wonder and awe that I had never experienced before. |
| *Response:* | { "response": "No" } |
| *System Prompt:* | Does the given prompt imply a travel experience, whether it is in the past, present, or future? Respond with a Yes or a No. Only respond as a JSON object. For example, { "response": "Yes" }. |
| *Prompt:* | As I soared above the Serengeti National Park in a hot air balloon, I felt a sense of wonder and awe that I had never experienced before. |
| *Response:* | { "response": "Yes" } |
| *System Prompt:* | Analyze the sentence and identify all SVO phrases. Replace any pronouns (e.g. "he" or "she" or "they" or "I" or "you") with proper nouns. Respond with ONLY a JSON object with only three fields: subject, object, and action. For example, if I say "Ben is so happy - he bought a boat and is going sailing", you should reply with { "subject": "Ben", "object": "a boat", "action": "bought" } |
| *Prompt:* | As I soared above the Serengeti National Park in a hot air balloon, I felt a sense of wonder and awe that I had never experienced before. |
| *Response:* | { |
| |    "subject": "I",<br>   "object": "a sense of wonder and awe",<br>   "action": "felt"<br>} |
| *System Prompt:* | Does this prompt have multiple subjects or objects? If a pronoun relates to a noun in the sentence then it is considered a single subject. Respond with only a Yes or No. Only respond as a JSON object. For example, { "response": "Yes" }. |
| *Prompt:* | The sun was rising over the savannah, casting a golden glow over the grasslands and acacia trees. |
| *Response:* | { "response": "No" } |
| *System Prompt:* | Does the given prompt imply a travel experience, whether it is in the past, present, or future? Respond with a Yes or a No. Only respond as a JSON object. For example, { "response": "Yes" }. |
| *Prompt:* | The sun was rising over the savannah, casting a golden glow over the grasslands and acacia trees. |
| *Response:* | { "response": "Yes" } |
| *System Prompt:* | Analyze the sentence and identify all SVO phrases. Replace any pronouns (e.g. "he" or "she" or "they" or "I" or "you") with proper nouns. Respond with ONLY a JSON object with only three fields: subject, object, and action. For example, if I say "Ben is so happy - he bought a boat and is going sailing", you should reply with { "subject": "Ben", "object": "a boat", "action": "bought" } |
| *Prompt:* | The sun was rising over the savannah, casting a golden glow over the grasslands and acacia trees. |
| *Response:* | {<br>   "subject": "the sun",<br>   "object": "a golden glow",<br>   "action": "casting"<br>} |
| *System Prompt:* | Does this prompt have multiple subjects or objects? If a pronoun relates to a noun in the sentence then it is considered a single subject. Respond with only a Yes or No. Only respond as a JSON object. For example, { "response": "Yes" }. |
| *Prompt:* | Below me, the majestic animals of Africa roamed freely - giraffes, zebras, and lions all seemed to be going about their day as if I wasn't there at all. |
| *Response:* | { "response": "No" } |
| *System Prompt:* | Does the given prompt imply a travel experience, whether it is in the past, present, or future? Respond with a Yes or a No. Only respond as a JSON object. For example, { "response": "Yes" }. |
| *Prompt:* | Below me, the majestic animals of Africa roamed freely - giraffes, zebras, and lions all seemed to be going about their day as if I wasn't there at all. |
| *Response:* | { "response": "Yes" } |
| *System Prompt:* | Analyze the sentence and identify all SVO phrases. Replace any pronouns (e.g. "he" or "she" or "they" or "I" or "you") with proper nouns. Respond with ONLY a JSON object with only three fields: subject, object, and action. For example, if I say "Ben is so happy - he bought a boat and is going sailing", you should reply with { "subject": "Ben", "object": "a boat", "action": "bought" } |
| *Prompt:* | Below me, the majestic animals of Africa roamed freely - giraffes, zebras, and lions all seemed to be going about their day as if I wasn't there at all. |
| *Response:* | {<br>   "subject": "the majestic animals of Africa",<br>   "object": "their day",<br>   "action": "seemed"<br>} |
| *System Prompt:* | Does this prompt have multiple subjects or objects? If a pronoun relates to a noun in the sentence then it is considered a single subject. Respond with only a Yes or No. Only respond as a JSON object. For example, { "response": "Yes" }. |
| *Prompt:* | Our pilot expertly navigated the balloon through the wispy clouds, offering breathtaking views of |
| | the landscape below. |
| *Response:* | { "response": "No" } |
| | |
| *System Prompt:* | Does the given prompt imply a travel experience, whether it is in the past, present, or future? Respond with a Yes or a No. Only respond as a JSON object. For example, { "response": "Yes" }. |
| *Prompt:* | Our pilot expertly navigated the balloon through the wispy clouds, offering breathtaking views of the landscape below. |
| *Response:* | { "response": "Yes" } |
| | |
| *System Prompt:* | Analyze the sentence and identify all SVO phrases. Replace any pronouns (e.g. "he" or "she" or "they" or "I" or "you") with proper nouns. Respond with ONLY a JSON object with only three fields: subject, object, and action. For example, if I say "Ben is so happy - he bought a boat and is going sailing", you should reply with { "subject": "Ben", "object": "a boat", "action": "bought" } |
| *Prompt:* | Our pilot expertly navigated the balloon through the wispy clouds, offering breathtaking views of the landscape below. |
| *Response:* | {<br>  "subject": "the pilot",<br>  "object": "the balloon",<br>  "action": "navigated"<br>} |
| | |
| *System Prompt:* | Does this prompt have multiple subjects or objects? If a pronoun relates to a noun in the sentence then it is considered a single subject. Respond with only a Yes or No. Only respond as a JSON object. For example, { "response": "Yes" }. |
| *Prompt:* | As we floated above a herd of grazing wildebeest, I felt a sense of peace wash over me - this was truly a once-in-a-lifetime experience. |
| *Response:* | { "response": "No" } |
| | |
| *System Prompt:* | Does the given prompt imply a travel experience, whether it is in the past, present, or future? Respond with a Yes or a No. Only respond as a JSON object. For example, { "response": "Yes" }. |
| *Prompt:* | As we floated above a herd of grazing wildebeest, I felt a sense of peace wash over me - this was truly a once-in-a-lifetime experience. |
| *Response:* | { "response": "Yes" } |
| | |
| *System Prompt:* | Analyze the sentence and identify all SVO phrases. Replace any pronouns (e.g. "he" or "she" or "they" or "I" or "you") with proper nouns. Respond with ONLY a JSON object with only three fields: subject, object, and action. For example, if I say "Ben is so happy - he bought a boat and is going sailing", you should reply with { "subject": "Ben", "object": "a boat", "action": "bought" } |
| *Prompt:* | As we floated above a herd of grazing wildebeest, I felt a sense of peace wash over me - this was truly a once-in-a-lifetime experience. |
| *Response:* | {<br>  "subject": "I",<br>  "object": "a sense of peace",<br>  "action": "wash"<br>} |

**Listing A13.** Test 13.
Start: 2:53:35 PM          End: 2:54:01 PM

| | |
|---|---|
| *System Prompt:* | Is the following input multiple independent clauses? Respond with only a Yes or No. Only respond as a JSON object. For example, "Dillan ate an apple. He thought it tasted good. " { "response": "Yes" }. |
| *Prompt:* | As I walked through the narrow canyon known as the Siq, I couldn't help but feel a sense of excitement and wonder. The ancient city of Petra stretched out before me, its intricately carved facades and temples rising up from the sandstone cliffs like giants. I had always been fascinated by history, and to see this incredible site in person was truly a dream come true. As I explored the |
| | ruins, I felt a sense of connection to the people who had built this city over 2,000 years ago - their ingenuity and craftsmanship were still evident in every stone. |
| *Response:* | { "response": "Yes" } |
| *System Prompt:* | Split this paragraph into distinct sentences. Only respond as a JSON array For example if given the text "This is one sentence. This is another sentence", respond with [ "This is one sentence.", "This is another sentence." ]. |
| *Prompt:* | As I walked through the narrow canyon known as the Siq, I couldn't help but feel a sense of excitement and wonder. The ancient city of Petra stretched out before me, its intricately carved facades and temples rising up from the sandstone cliffs like giants. I had always been fascinated by history, and to see this incredible site in person was truly a dream come true. As I explored the ruins, I felt a sense of connection to the people who had built this city over 2,000 years ago - their ingenuity and craftsmanship were still evident in every stone. |
| *Response:* | [<br>  "As I walked through the narrow canyon known as the Siq, I couldn't help but feel a sense of excitement and wonder.",<br>  "The ancient city of Petra stretched out before me, its intricately carved facades and temples rising up from the sandstone cliffs like giants.",<br>  "I had always been fascinated by history, and to see this incredible site in person was truly a dream come true.",<br>  "As I explored the ruins, I felt a sense of connection to the people who had built this city over 2,000 years ago - their ingenuity and craftsmanship were still evident in every stone."<br>] |
| *System Prompt:* | Does this prompt have multiple subjects or objects? If a pronoun relates to a noun in the sentence then it is considered a single subject. Respond with only a Yes or No. Only respond as a JSON object. For example, { "response": "Yes" }. |
| *Prompt:* | As I walked through the narrow canyon known as the Siq, I couldn't help but feel a sense of excitement and wonder. |
| *Response:* | { "response": "No" } |
| *System Prompt:* | Does the given prompt imply a travel experience, whether it is in the past, present, or future? Respond with a Yes or a No. Only respond as a JSON object. For example, { "response": "Yes" }. |
| *Prompt:* | As I walked through the narrow canyon known as the Siq, I couldn't help but feel a sense of excitement and wonder. |
| *Response:* | { "response": "Yes" } |
| *System Prompt:* | Analyze the sentence and identify all SVO phrases. Replace any pronouns (e.g. "he" or "she" or "they" or "I" or "you") with proper nouns. Respond with ONLY a JSON object with only three fields: subject, object, and action. For example, if I say "Ben is so happy - he bought a boat and is going sailing", you should reply with { "subject": "Ben", "object": "a boat", "action": "bought" } |
| *Prompt:* | As I walked through the narrow canyon known as the Siq, I couldn't help but feel a sense of excitement and wonder. |
| *Response:* | {<br>  "subject": "I",<br>  "object": "excitement and wonder",<br>  "action": "feel"<br>} |
| *System Prompt:* | Does this prompt have multiple subjects or objects? If a pronoun relates to a noun in the sentence then it is considered a single subject. Respond with only a Yes or No. Only respond as a JSON object. For example, { "response": "Yes" }. |
| *Prompt:* | The ancient city of Petra stretched out before me, its intricately carved facades and temples rising up from the sandstone cliffs like giants. |
| *Response:* | { "response": "No" } |
| *System Prompt:* | Does the given prompt imply a travel experience, whether it is in the past, present, or future? Respond with a Yes or a No. Only respond as a JSON object. For example, { "response": "Yes" }. |
| *Prompt:* | The ancient city of Petra stretched out before me, its intricately carved facades and temples rising up from the sandstone cliffs like giants. |
| *Response:* | { "response": "Yes" } |
| | |
| *System Prompt:* | Analyze the sentence and identify all SVO phrases. Replace any pronouns (e.g. "he" or "she" or "they" or "I" or "you") with proper nouns. Respond with ONLY a JSON object with only three fields: subject, object, and action. For example, if I say "Ben is so happy - he bought a boat and is going sailing", you should reply with { "subject": "Ben", "object": "a boat", "action": "bought" } |
| *Prompt:* | The ancient city of Petra stretched out before me, its intricately carved facades and temples rising up from the sandstone cliffs like giants. |
| *Response:* | {<br>  "subject": "Petra",<br>  "object": "the sandstone cliffs",<br>  "action": "rising"<br>} |
| | |
| *System Prompt:* | Does this prompt have multiple subjects or objects? If a pronoun relates to a noun in the sentence then it is considered a single subject. Respond with only a Yes or No. Only respond as a JSON object. For example, { "response": "Yes" }. |
| *Prompt:* | I had always been fascinated by history, and to see this incredible site in person was truly a dream come true. |
| *Response:* | { "response": "No" } |
| | |
| *System Prompt:* | Does the given prompt imply a travel experience, whether it is in the past, present, or future? Respond with a Yes or a No. Only respond as a JSON object. For example, { "response": "Yes" }. |
| *Prompt:* | I had always been fascinated by history, and to see this incredible site in person was truly a dream come true. |
| *Response:* | { "response": "Yes" } |
| | |
| *System Prompt:* | Analyze the sentence and identify all SVO phrases. Replace any pronouns (e.g. "he" or "she" or "they" or "I" or "you") with proper nouns. Respond with ONLY a JSON object with only three fields: subject, object, and action. For example, if I say "Ben is so happy - he bought a boat and is going sailing", you should reply with { "subject": "Ben", "object": "a boat", "action": "bought" } |
| *Prompt:* | I had always been fascinated by history, and to see this incredible site in person was truly a dream come true. |
| *Response:* | {<br>  "subject": "the speaker",<br>  "object": "this incredible site",<br>  "action": "see"<br>} |
| | |
| *System Prompt:* | Does this prompt have multiple subjects or objects? If a pronoun relates to a noun in the sentence then it is considered a single subject. Respond with only a Yes or No. Only respond as a JSON object. For example, { "response": "Yes" }. |
| *Prompt:* | As I explored the ruins, I felt a sense of connection to the people who had built this city over 2,000 years ago - their ingenuity and craftsmanship were still evident in every stone. |
| *Response:* | { "response": "No" } |
| | |
| *System Prompt:* | Does the given prompt imply a travel experience, whether it is in the past, present, or future? Respond with a Yes or a No. Only respond as a JSON object. For example, { "response": "Yes" }. |
| *Prompt:* | As I explored the ruins, I felt a sense of connection to the people who had built this city over 2,000 years ago - their ingenuity and craftsmanship were still evident in every stone. |
| *Response:* | { "response": "Yes" } |

*System Prompt:* Analyze the sentence and identify all SVO phrases. Replace any pronouns (e.g. "he" or "she" or "they" or "I" or "you") with proper nouns. Respond with ONLY a JSON object with only three fields: subject, object, and action. For example, if I say "Ben is so happy - he bought a boat and is going sailing", you should reply with { "subject": "Ben", "object": "a boat", "action": "bought" }

*Prompt:* As I explored the ruins, I felt a sense of connection to the people who had built this city over 2,000 years ago - their ingenuity and craftsmanship were still evident in every stone.

*Response:*
```
{
  "subject": "I",
  "object": "the people who had built this city",
  "action": "felt"
}
```

**Listing A14.** Test 14.

Start: 2:57:06 PM          End: 2:57:28 PM

*System Prompt:* Is the following input multiple independent clauses? Respond with only a Yes or No. Only respond as a JSON object. For example, "Dillan ate an apple. He thought it tasted good. " { "response": "Yes" }.

*Prompt:* I had always loved Italian food, but nothing could have prepared me for the culinary experience that awaited me in Tuscany. As I walked through the rolling hills and vineyards of this picturesque region, I was invited to join a cooking class at a local villa. Under the guidance of our chef, we learned how to prepare traditional Tuscan dishes using fresh ingredients from the surrounding countryside. The aromas wafting from the kitchen were incredible - pasta sauces simmered on the stovetop, while freshly baked bread and pizza dough rose in the oven. As we sat down to enjoy our creations, I felt a sense of satisfaction and joy that came from sharing a meal with new friends.

*Response:* { "response": "Yes" }

*System Prompt:* Split this paragraph into distinct sentences. Only respond as a JSON array For example if given the text "This is one sentence. This is another sentence", respond with [ "This is one sentence.", "This is another sentence." ].

*Prompt:* I had always loved Italian food, but nothing could have prepared me for the culinary experience that awaited me in Tuscany. As I walked through the rolling hills and vineyards of this picturesque region, I was invited to join a cooking class at a local villa. Under the guidance of our chef, we learned how to prepare traditional Tuscan dishes using fresh ingredients from the surrounding countryside. The aromas wafting from the kitchen were incredible - pasta sauces simmered on the stovetop, while freshly baked bread and pizza dough rose in the oven. As we sat down to enjoy our creations, I felt a sense of satisfaction and joy that came from sharing a meal with new friends.

*Response:*
```
[
   "I had always loved Italian food, but nothing could have prepared me for the culinary experience that awaited me in Tuscany.",
   "As I walked through the rolling hills and vineyards of this picturesque region, I was invited to join a cooking class at a local villa.",
   "Under the guidance of our chef, we learned how to prepare traditional Tuscan dishes using fresh ingredients from the surrounding countryside.",
   "The aromas wafting from the kitchen were incredible - pasta sauces simmered on the stovetop, while freshly baked bread and pizza dough rose in the oven.",
   "As we sat down to enjoy our creations, I felt a sense of satisfaction and joy that came from sharing a meal with new friends."
]
```

*System Prompt:* Does this prompt have multiple subjects or objects? If a pronoun relates to a noun in the sentence then it is considered a single subject. Respond with only a Yes or No. Only respond as a JSON object. For example, { "response": "Yes" }.

*Prompt:* I had always loved Italian food, but nothing could have prepared me for the culinary experience that awaited me in Tuscany.

*Response:* { "response": "No" }

| | |
|---|---|
| *System Prompt:* | Does the given prompt imply a travel experience, whether it is in the past, present, or future? Respond with a Yes or a No. Only respond as a JSON object. For example, { "response": "Yes" }. |
| *Prompt:* | I had always loved Italian food, but nothing could have prepared me for the culinary experience that awaited me in Tuscany. |
| *Response:* | { "response": "Yes" } |
| *System Prompt:* | Analyze the sentence and identify all SVO phrases. Replace any pronouns (e.g. "he" or "she" or "they" or "I" or "you") with proper nouns. Respond with ONLY a JSON object with only three fields: subject, object, and action. For example, if I say "Ben is so happy - he bought a boat and is going sailing", you should reply with { "subject": "Ben", "object": "a boat", "action": "bought" } |
| *Prompt:* | I had always loved Italian food, but nothing could have prepared me for the culinary experience that awaited me in Tuscany. |
| *Response:* | {<br>  "subject": "me",<br>  "object": "the culinary experience",<br>  "action": "awaited"<br>} |
| *System Prompt:* | Does this prompt have multiple subjects or objects? If a pronoun relates to a noun in the sentence then it is considered a single subject. Respond with only a Yes or No. Only respond as a JSON object. For example, { "response": "Yes" }. |
| *Prompt:* | As I walked through the rolling hills and vineyards of this picturesque region, I was invited to join a cooking class at a local villa. |
| *Response:* | { "response": "No" } |
| *System Prompt:* | Does the given prompt imply a travel experience, whether it is in the past, present, or future? Respond with a Yes or a No. Only respond as a JSON object. For example, { "response": "Yes" }. |
| *Prompt:* | As I walked through the rolling hills and vineyards of this picturesque region, I was invited to join a cooking class at a local villa. |
| *Response:* | { "response": "Yes" } |
| *System Prompt:* | Analyze the sentence and identify all SVO phrases. Replace any pronouns (e.g. "he" or "she" or "they" or "I" or "you") with proper nouns. Respond with ONLY a JSON object with only three fields: subject, object, and action. For example, if I say "Ben is so happy - he bought a boat and is going sailing", you should reply with { "subject": "Ben", "object": "a boat", "action": "bought" } |
| *Prompt:* | As I walked through the rolling hills and vineyards of this picturesque region, I was invited to join a cooking class at a local villa. |
| *Response:* | {<br>  "subject": "I",<br>  "object": "a cooking class",<br>  "action": "invited"<br>} |
| *System Prompt:* | Does this prompt have multiple subjects or objects? If a pronoun relates to a noun in the sentence then it is considered a single subject. Respond with only a Yes or No. Only respond as a JSON object. For example, { "response": "Yes" }. |
| *Prompt:* | Under the guidance of our chef, we learned how to prepare traditional Tuscan dishes using fresh ingredients from the surrounding countryside. |
| *Response:* | { "response": "No" } |
| *System Prompt:* | Does the given prompt imply a travel experience, whether it is in the past, present, or future? Respond with a Yes or a No. Only respond as a JSON object. For example, { "response": "Yes" }. |
| *Prompt:* | Under the guidance of our chef, we learned how to prepare traditional Tuscan dishes using fresh ingredients from the surrounding countryside. |
| *Response:* | { "response": "No" } |
| *System Prompt:* | Does this prompt have multiple subjects or objects? If a pronoun relates to a noun in the sentence then it is considered a single subject. Respond with only a Yes or No. Only respond as a JSON object. For example, { "response": "Yes" }. |
| *Prompt:* | The aromas wafting from the kitchen were incredible - pasta sauces simmered on the stovetop, while freshly baked bread and pizza dough rose in the oven. |
| *Response:* | { "response": "No" } |
| | |
| *System Prompt:* | Does the given prompt imply a travel experience, whether it is in the past, present, or future? Respond with a Yes or a No. Only respond as a JSON object. For example, { "response": "Yes" }. |
| *Prompt:* | The aromas wafting from the kitchen were incredible - pasta sauces simmered on the stovetop, while freshly baked bread and pizza dough rose in the oven. |
| *Response:* | { "response": "No" } |
| | |
| *System Prompt:* | Does this prompt have multiple subjects or objects? If a pronoun relates to a noun in the sentence then it is considered a single subject. Respond with only a Yes or No. Only respond as a JSON object. For example, { "response": "Yes" }. |
| *Prompt:* | As we sat down to enjoy our creations, I felt a sense of satisfaction and joy that came from sharing a meal with new friends. |
| *Response:* | { "response": "No" } |
| | |
| *System Prompt:* | Does the given prompt imply a travel experience, whether it is in the past, present, or future? Respond with a Yes or a No. Only respond as a JSON object. For example, { "response": "Yes" }. |
| *Prompt:* | As we sat down to enjoy our creations, I felt a sense of satisfaction and joy that came from sharing a meal with new friends. |
| *Response:* | { "response": "No" } |

**Listing A15.** Test 15.
Start: 2:58:21 PM        End: 2:58:37 PM

| | |
|---|---|
| *System Prompt:* | Is the following input multiple independent clauses? Respond with only a Yes or No. Only respond as a JSON object. For example, "Dillan ate an apple. He thought it tasted good. " { "response": "Yes" }. |
| *Prompt:* | As I slipped into the crystal-clear waters of Bora Bora's lagoon, I was immediately struck by the vibrant colors and incredible marine life that surrounded me. The sea turtles swam alongside me, their gentle eyes seeming to regard me with a quiet curiosity. As we explored the underwater world together, I felt a sense of connection to these magnificent creatures - they were so much more than just animals; they were ambassadors for the beauty and wonder of our planet. |
| *Response:* | { "response": "Yes" } |
| | |
| *System Prompt:* | Split this paragraph into distinct sentences. Only respond as a JSON array For example if given the text "This is one sentence. This is another sentence", respond with [ "This is one sentence.", "This is another sentence." ]. |
| *Prompt:* | As I slipped into the crystal-clear waters of Bora Bora's lagoon, I was immediately struck by the vibrant colors and incredible marine life that surrounded me. The sea turtles swam alongside me, their gentle eyes seeming to regard me with a quiet curiosity. As we explored the underwater world together, I felt a sense of connection to these magnificent creatures - they were so much more than just animals; they were ambassadors for the beauty and wonder of our planet. |
| *Response:* | [<br>  "As I slipped into the crystal-clear waters of Bora Bora's lagoon, I was immediately struck by the vibrant colors and incredible marine life that surrounded me.",<br>  "The sea turtles swam alongside me, their gentle eyes seeming to regard me with a quiet curiosity.",<br>  "As we explored the underwater world together, I felt a sense of connection to these magnificent creatures - they were so much more than just animals; they were ambassadors for the beauty and wonder of our planet." |

]

*System Prompt:*   Does this prompt have multiple subjects or objects? If a pronoun relates to a noun in the sentence then it is considered a single subject. Respond with only a Yes or No. Only respond as a JSON object. For example, { "response": "Yes" }.
*Prompt:*   As I slipped into the crystal-clear waters of Bora Bora's lagoon, I was immediately struck by the vibrant colors and incredible marine life that surrounded me.
*Response:*   { "response": "No" }

*System Prompt:*   Does the given prompt imply a travel experience, whether it is in the past, present, or future? Respond with a Yes or a No. Only respond as a JSON object. For example, { "response": "Yes" }.
*Prompt:*   As I slipped into the crystal-clear waters of Bora Bora's lagoon, I was immediately struck by the vibrant colors and incredible marine life that surrounded me.
*Response:*   { "response": "Yes" }

*System Prompt:*   Analyze the sentence and identify all SVO phrases. Replace any pronouns (e.g. "he" or "she" or "they" or "I" or "you") with proper nouns. Respond with ONLY a JSON object with only three fields: subject, object, and action. For example, if I say "Ben is so happy - he bought a boat and is going sailing", you should reply with { "subject": "Ben", "object": "a boat", "action": "bought" }
*Prompt:*   As I slipped into the crystal-clear waters of Bora Bora's lagoon, I was immediately struck by the vibrant colors and incredible marine life that surrounded me.
*Response:*
```
{
   "subject": "I",
   "object": "vibrant colors and incredible marine life",
   "action": "struck"
}
```

*System Prompt:*   Does this prompt have multiple subjects or objects? If a pronoun relates to a noun in the sentence then it is considered a single subject. Respond with only a Yes or No. Only respond as a JSON object. For example, { "response": "Yes" }.
*Prompt:*   The sea turtles swam alongside me, their gentle eyes seeming to regard me with a quiet curiosity.
*Response:*   { "response": "No" }

*System Prompt:*   Does the given prompt imply a travel experience, whether it is in the past, present, or future? Respond with a Yes or a No. Only respond as a JSON object. For example, { "response": "Yes" }.
*Prompt:*   The sea turtles swam alongside me, their gentle eyes seeming to regard me with a quiet curiosity.
*Response:*   { "response": "Yes" }

*System Prompt:*   Analyze the sentence and identify all SVO phrases. Replace any pronouns (e.g. "he" or "she" or "they" or "I" or "you") with proper nouns. Respond with ONLY a JSON object with only three fields: subject, object, and action. For example, if I say "Ben is so happy - he bought a boat and is going sailing", you should reply with { "subject": "Ben", "object": "a boat", "action": "bought" }
*Prompt:*   The sea turtles swam alongside me, their gentle eyes seeming to regard me with a quiet curiosity.
*Response:*
```
{
   "subject": "the sea turtles",
   "object": "me",
   "action": "regard"
}
```

*System Prompt:*   Does this prompt have multiple subjects or objects? If a pronoun relates to a noun in the sentence then it is considered a single subject. Respond with only a Yes or No. Only respond as a JSON object. For example, { "response": "Yes" }.
*Prompt:*   As we explored the underwater world together, I felt a sense of connection to these magnificent creatures - they were so much more than just animals; they were ambassadors for the beauty and wonder of our planet.
*Response:*   { "response": "No" }

| | |
|---|---|
| *System Prompt:* | Does the given prompt imply a travel experience, whether it is in the past, present, or future? Respond with a Yes or a No. Only respond as a JSON object. For example, { "response": "Yes" }. |
| *Prompt:* | As we explored the underwater world together, I felt a sense of connection to these magnificent creatures - they were so much more than just animals; they were ambassadors for the beauty and wonder of our planet. |
| *Response:* | { "response": "No" } |

**Listing A16.** Test 16.
Start: 2:59:36 PM          End: 2:59:59 PM

| | |
|---|---|
| *System Prompt:* | Is the following input multiple independent clauses? Respond with only a Yes or No. Only respond as a JSON object. For example, "Dillan ate an apple. He thought it tasted good. " { "response": "Yes" }. |
| *Prompt:* | As I stood before the majestic Taj Mahal at sunrise, I felt a sense of awe and reverence wash over me. The intricate marble carvings and ornate details seemed to glow with an otherworldly light in the early morning sun. As I walked through the gardens and reflecting pools surrounding this incredible monument, I couldn't help but feel a sense of wonder at the love story that had inspired its creation - Shah Jahan's devotion to his beloved wife was still palpable in every stone. |
| *Response:* | { "response": "Yes" } |
| *System Prompt:* | Split this paragraph into distinct sentences. Only respond as a JSON array For example if given the text "This is one sentence. This is another sentence", respond with [ "This is one sentence.", "This is another sentence." ]. |
| *Prompt:* | As I stood before the majestic Taj Mahal at sunrise, I felt a sense of awe and reverence wash over me. The intricate marble carvings and ornate details seemed to glow with an otherworldly light in the early morning sun. As I walked through the gardens and reflecting pools surrounding this incredible monument, I couldn't help but feel a sense of wonder at the love story that had inspired its creation - Shah Jahan's devotion to his beloved wife was still palpable in every stone. |
| *Response:* | [<br>  "As I stood before the majestic Taj Mahal at sunrise, I felt a sense of awe and reverence wash over me.",<br>  "The intricate marble carvings and ornate details seemed to glow with an otherworldly light in the early morning sun.",<br>  "As I walked through the gardens and reflecting pools surrounding this incredible monument,",<br>  "I couldn't help but feel a sense of wonder at the love story that had inspired its creation - Shah Jahan's devotion to his beloved wife was still palpable in every stone."<br>] |
| *System Prompt:* | Does this prompt have multiple subjects or objects? If a pronoun relates to a noun in the sentence then it is considered a single subject. Respond with only a Yes or No. Only respond as a JSON object. For example, { "response": "Yes" }. |
| *Prompt:* | As I stood before the majestic Taj Mahal at sunrise, I felt a sense of awe and reverence wash over me. |
| *Response:* | { "response": "No" } |
| *System Prompt:* | Does the given prompt imply a travel experience, whether it is in the past, present, or future? Respond with a Yes or a No. Only respond as a JSON object. For example, { "response": "Yes" }. |
| *Prompt:* | As I stood before the majestic Taj Mahal at sunrise, I felt a sense of awe and reverence wash over me. |
| *Response:* | { "response": "Yes" } |
| *System Prompt:* | Analyze the sentence and identify all SVO phrases. Replace any pronouns (e.g. "he" or "she" or "they" or "I" or "you") with proper nouns. Respond with ONLY a JSON object with only three fields: subject, object, and action. For example, if I say "Ben is so happy - he bought a boat and is going sailing", you should reply with { "subject": "Ben", "object": "a boat", "action": "bought" } |
| *Prompt:* | As I stood before the majestic Taj Mahal at sunrise, I felt a sense of awe and reverence wash over me. |
| *Response:* | {<br>  "subject": "I",<br>  "object": "awe and reverence",<br>  "action": "wash"<br>} |
| *System Prompt:* | Does this prompt have multiple subjects or objects? If a pronoun relates to a noun in the sentence then it is considered a single subject. Respond with only a Yes or No. Only respond as a JSON object. For example, { "response": "Yes" }. |
| *Prompt:* | The intricate marble carvings and ornate details seemed to glow with an otherworldly light in the early morning sun. |
| *Response:* | { "response": "No" } |
| *System Prompt:* | Does the given prompt imply a travel experience, whether it is in the past, present, or future? Respond with a Yes or a No. Only respond as a JSON object. For example, { "response": "Yes" }. |
| *Prompt:* | The intricate marble carvings and ornate details seemed to glow with an otherworldly light in the early morning sun. |
| *Response:* | { "response": "Yes" } |
| *System Prompt:* | Analyze the sentence and identify all SVO phrases. Replace any pronouns (e.g. "he" or "she" or "they" or "I" or "you") with proper nouns. Respond with ONLY a JSON object with only three fields: subject, object, and action. For example, if I say "Ben is so happy - he bought a boat and is going sailing", you should reply with { "subject": "Ben", "object": "a boat", "action": "bought" } |
| *Prompt:* | The intricate marble carvings and ornate details seemed to glow with an otherworldly light in the early morning sun. |
| *Response:* | {<br>  "subject": "the intricate marble carvings and ornate details",<br>  "object": "an otherworldly light",<br>  "action": "seemed"<br>} |
| *System Prompt:* | Does this prompt have multiple subjects or objects? If a pronoun relates to a noun in the sentence then it is considered a single subject. Respond with only a Yes or No. Only respond as a JSON object. For example, { "response": "Yes" }. |
| *Prompt:* | As I walked through the gardens and reflecting pools surrounding this incredible monument, |
| *Response:* | { "response": "No" } |
| *System Prompt:* | Does the given prompt imply a travel experience, whether it is in the past, present, or future? Respond with a Yes or a No. Only respond as a JSON object. For example, { "response": "Yes" }. |
| *Prompt:* | As I walked through the gardens and reflecting pools surrounding this incredible monument, |
| *Response:* | { "response": "Yes" } |
| *System Prompt:* | Analyze the sentence and identify all SVO phrases. Replace any pronouns (e.g. "he" or "she" or "they" or "I" or "you") with proper nouns. Respond with ONLY a JSON object with only three fields: subject, object, and action. For example, if I say "Ben is so happy - he bought a boat and is going sailing", you should reply with { "subject": "Ben", "object": "a boat", "action": "bought" } |
| *Prompt:* | As I walked through the gardens and reflecting pools surrounding this incredible monument, |
| *Response:* | {<br>  "subject": "I",<br>  "object": "the gardens and reflecting pools",<br>  "action": "walked"<br>} |
| *System Prompt:* | Does this prompt have multiple subjects or objects? If a pronoun relates to a noun in the sentence |
| | then it is considered a single subject. Respond with only a Yes or No. Only respond as a JSON object. For example, { "response": "Yes" }. |
| *Prompt:* | I couldn't help but feel a sense of wonder at the love story that had inspired its creation - Shah Jahan's devotion to his beloved wife was still palpable in every stone. |
| *Response:* | { "response": "No" } |
| | |
| *System Prompt:* | Does the given prompt imply a travel experience, whether it is in the past, present, or future? Respond with a Yes or a No. Only respond as a JSON object. For example, { "response": "Yes" }. |
| *Prompt:* | I couldn't help but feel a sense of wonder at the love story that had inspired its creation - Shah Jahan's devotion to his beloved wife was still palpable in every stone. |
| *Response:* | { "response": "No" } |

**Listing 17.** Test 17.
Start: 3:02:13 PM                End: 3:02:32 PM

| | |
|---|---|
| *System Prompt:* | Is the following input multiple independent clauses? Respond with only a Yes or No. Only respond as a JSON object. For example, "Dillan ate an apple. He thought it tasted good. " { "response": "Yes" }. |
| *Prompt:* | As I drove through the stunning landscapes of New Zealand's South Island, I felt a sense of freedom and adventure that I had never experienced before. The rugged mountains, pristine lakes, and picturesque villages all seemed to unfold like a tapestry before me - each new turn in the road revealed another breathtaking view or hidden gem. As we stopped at a local café for a cup of coffee and a slice of homemade pie, I felt grateful for this chance to explore one of the most beautiful countries on earth. |
| *Response:* | { "response": "Yes" } |
| | |
| *System Prompt:* | Split this paragraph into distinct sentences. Only respond as a JSON array For example if given the text "This is one sentence. This is another sentence", respond with [ "This is one sentence.", "This is another sentence." ]. |
| *Prompt:* | As I drove through the stunning landscapes of New Zealand's South Island, I felt a sense of freedom and adventure that I had never experienced before. The rugged mountains, pristine lakes, and picturesque villages all seemed to unfold like a tapestry before me - each new turn in the road revealed another breathtaking view or hidden gem. As we stopped at a local café for a cup of coffee and a slice of homemade pie, I felt grateful for this chance to explore one of the most beautiful countries on earth. |
| *Response:* | [<br>  "As I drove through the stunning landscapes of New Zealand's South Island, I felt a sense of freedom and adventure that I had never experienced before.",<br>  "The rugged mountains, pristine lakes, and picturesque villages all seemed to unfold like a tapestry before me - each new turn in the road revealed another breathtaking view or hidden gem.",<br>  "As we stopped at a local café for a cup of coffee and a slice of homemade pie, I felt grateful for this chance to explore one of the most beautiful countries on earth."<br>] |
| | |
| *System Prompt:* | Does this prompt have multiple subjects or objects? If a pronoun relates to a noun in the sentence then it is considered a single subject. Respond with only a Yes or No. Only respond as a JSON object. For example, { "response": "Yes" }. |
| *Prompt:* | As I drove through the stunning landscapes of New Zealand's South Island, I felt a sense of freedom and adventure that I had never experienced before. |
| *Response:* | { "response": "No" } |
| | |
| *System Prompt:* | Does the given prompt imply a travel experience, whether it is in the past, present, or future? Respond with a Yes or a No. Only respond as a JSON object. For example, { "response": "Yes" }. |
| *Prompt:* | As I drove through the stunning landscapes of New Zealand's South Island, I felt a sense of freedom and adventure that I had never experienced before. |
| *Response:* | { "response": "Yes" } |
| *System Prompt:* | Analyze the sentence and identify all SVO phrases. Replace any pronouns (e.g. "he" or "she" or "they" or "I" or "you") with proper nouns. Respond with ONLY a JSON object with only three fields: subject, object, and action. For example, if I say "Ben is so happy - he bought a boat and is going sailing", you should reply with { "subject": "Ben", "object": "a boat", "action": "bought" } |
| *Prompt:* | As I drove through the stunning landscapes of New Zealand's South Island, I felt a sense of freedom and adventure that I had never experienced before. |
| *Response:* | {<br>  "subject": "I",<br>  "object": "a sense of freedom and adventure",<br>  "action": "felt"<br>} |
| *System Prompt:* | Does this prompt have multiple subjects or objects? If a pronoun relates to a noun in the sentence then it is considered a single subject. Respond with only a Yes or No. Only respond as a JSON object. For example, { "response": "Yes" }. |
| *Prompt:* | The rugged mountains, pristine lakes, and picturesque villages all seemed to unfold like a tapestry before me - each new turn in the road revealed another breathtaking view or hidden gem. |
| *Response:* | { "response": "No" } |
| *System Prompt:* | Does the given prompt imply a travel experience, whether it is in the past, present, or future? Respond with a Yes or a No. Only respond as a JSON object. For example, { "response": "Yes" }. |
| *Prompt:* | The rugged mountains, pristine lakes, and picturesque villages all seemed to unfold like a tapestry before me - each new turn in the road revealed another breathtaking view or hidden gem. |
| *Response:* | { "response": "Yes" } |
| *System Prompt:* | Analyze the sentence and identify all SVO phrases. Replace any pronouns (e.g. "he" or "she" or "they" or "I" or "you") with proper nouns. Respond with ONLY a JSON object with only three fields: subject, object, and action. For example, if I say "Ben is so happy - he bought a boat and is going sailing", you should reply with { "subject": "Ben", "object": "a boat", "action": "bought" } |
| *Prompt:* | The rugged mountains, pristine lakes, and picturesque villages all seemed to unfold like a tapestry before me - each new turn in the road revealed another breathtaking view or hidden gem. |
| *Response:* | {<br>  "subject": "me",<br>  "object": "a tapestry",<br>  "action": "unfold"<br>} |
| *System Prompt:* | Does this prompt have multiple subjects or objects? If a pronoun relates to a noun in the sentence then it is considered a single subject. Respond with only a Yes or No. Only respond as a JSON object. For example, { "response": "Yes" }. |
| *Prompt:* | As we stopped at a local café for a cup of coffee and a slice of homemade pie, I felt grateful for this chance to explore one of the most beautiful countries on earth. |
| *Response:* | { "response": "No" } |
| *System Prompt:* | Does the given prompt imply a travel experience, whether it is in the past, present, or future? Respond with a Yes or a No. Only respond as a JSON object. For example, { "response": "Yes" }. |
| *Prompt:* | As we stopped at a local café for a cup of coffee and a slice of homemade pie, I felt grateful for this chance to explore one of the most beautiful countries on earth. |
| *Response:* | { "response": "Yes" } |
| *System Prompt:* | Analyze the sentence and identify all SVO phrases. Replace any pronouns (e.g. "he" or "she" or "they" or "I" or "you") with proper nouns. Respond with ONLY a JSON object with only three fields: subject, object, and action. For example, if I say "Ben is so happy - he bought a boat and is going sailing", you should reply with { "subject": "Ben", "object": "a boat", "action": "bought" } |
| *Prompt:* | As we stopped at a local café for a cup of coffee and a slice of homemade pie, I felt grateful for this chance to explore one of the most beautiful countries on earth. |
| *Response:* | {<br>  "subject": "I",<br>  "object": "a chance",<br>  "action": "felt"<br>} |

**Listing A18.** Test 18.
Start: 9:06:59 AM        End: 9:07:32 AM

| | |
|---|---|
| *System Prompt:* | Is the following input multiple independent clauses? Respond with only a Yes or No. Only respond as a JSON object. For example, "Dillan ate an apple. He thought it tasted good. " { "response": "Yes" }. |
| *Prompt:* | As I walked through the crumbling temples and ruins of Angkor Wat, I couldn't help but feel a sense of wonder at the ingenuity and craftsmanship of the people who had built this incredible site over 1,000 years ago. The intricate carvings and bas-reliefs seemed to tell stories of myth and legend, while the imposing stone structures rose up from the jungle like giants. As I explored the ancient city, I felt a sense of connection to the people who had once called this place home - their history and culture still lingered in every stone. |
| *Response:* | { "response": "Yes" } |
| | |
| *System Prompt:* | Split this paragraph into distinct sentences. Only respond as a JSON array For example if given the text "This is one sentence. This is another sentence", respond with [ "This is one sentence.", "This is another sentence." ]. |
| *Prompt:* | As I walked through the crumbling temples and ruins of Angkor Wat, I couldn't help but feel a sense of wonder at the ingenuity and craftsmanship of the people who had built this incredible site over 1,000 years ago. The intricate carvings and bas-reliefs seemed to tell stories of myth and legend, while the imposing stone structures rose up from the jungle like giants. As I explored the ancient city, I felt a sense of connection to the people who had once called this place home - their history and culture still lingered in every stone. |
| *Response:* | [<br>  "As I walked through the crumbling temples and ruins of Angkor Wat,",<br>  "I couldn't help but feel a sense of wonder at the ingenuity and craftsmanship of the people who had built this incredible site over 1,000 years ago.",<br>  "The intricate carvings and bas-reliefs seemed to tell stories of myth and legend,",<br>  "while the imposing stone structures rose up from the jungle like giants.",<br>  "As I explored the ancient city,",<br>  "I felt a sense of connection to the people who had once called this place home - their history and culture still lingered in every stone."<br>] |
| | |
| *System Prompt:* | Does this prompt have multiple subjects or objects? If a pronoun relates to a noun in the sentence then it is considered a single subject. Respond with only a Yes or No. Only respond as a JSON object. For example, { "response": "Yes" }. |
| *Prompt:* | As I walked through the crumbling temples and ruins of Angkor Wat, |
| *Response:* | { "response": "No" } |
| | |
| *System Prompt:* | Does the given prompt imply a travel experience, whether it is in the past, present, or future? Respond with a Yes or a No. Only respond as a JSON object. For example, { "response": "Yes" }. |
| *Prompt:* | As I walked through the crumbling temples and ruins of Angkor Wat, |
| *Response:* | { "response": "Yes" } |
| | |
| *System Prompt:* | Analyze the sentence and identify all SVO phrases. Replace any pronouns (e.g. "he" or "she" or "they" or "I" or "you") with proper nouns. Respond with ONLY a JSON object with only three fields: subject, object, and action. For example, if I say "Ben is so happy - he bought a boat and is |
| | going sailing", you should reply with { "subject": "Ben", "object": "a boat", "action": "bought" } |
| *Prompt:* | As I walked through the crumbling temples and ruins of Angkor Wat, |
| *Response:* | { "subject": "I", "object": "the crumbling temples and ruins of Angkor Wat", "action": "walked" } |
| *System Prompt:* | Does this prompt have multiple subjects or objects? If a pronoun relates to a noun in the sentence then it is considered a single subject. Respond with only a Yes or No. Only respond as a JSON object. For example, { "response": "Yes" }. |
| *Prompt:* | I couldn't help but feel a sense of wonder at the ingenuity and craftsmanship of the people who had built this incredible site over 1,000 years ago. |
| *Response:* | { "response": "No" } |
| *System Prompt:* | Does the given prompt imply a travel experience, whether it is in the past, present, or future? Respond with a Yes or a No. Only respond as a JSON object. For example, { "response": "Yes" }. |
| *Prompt:* | I couldn't help but feel a sense of wonder at the ingenuity and craftsmanship of the people who had built this incredible site over 1,000 years ago. |
| *Response:* | { "response": "Yes" } |
| *System Prompt:* | Analyze the sentence and identify all SVO phrases. Replace any pronouns (e.g. "he" or "she" or "they" or "I" or "you") with proper nouns. Respond with ONLY a JSON object with only three fields: subject, object, and action. For example, if I say "Ben is so happy - he bought a boat and is going sailing", you should reply with { "subject": "Ben", "object": "a boat", "action": "bought" } |
| *Prompt:* | I couldn't help but feel a sense of wonder at the ingenuity and craftsmanship of the people who had built this incredible site over 1,000 years ago. |
| *Response:* | {<br>  "subject": "the people",<br>  "object": "this incredible site",<br>  "action": "built"<br>} |
| *System Prompt:* | Does this prompt have multiple subjects or objects? If a pronoun relates to a noun in the sentence then it is considered a single subject. Respond with only a Yes or No. Only respond as a JSON object. For example, { "response": "Yes" }. |
| *Prompt:* | The intricate carvings and bas-reliefs seemed to tell stories of myth and legend, |
| *Response:* | { "response": "No" } |
| *System Prompt:* | Does the given prompt imply a travel experience, whether it is in the past, present, or future? Respond with a Yes or a No. Only respond as a JSON object. For example, { "response": "Yes" }. |
| *Prompt:* | The intricate carvings and bas-reliefs seemed to tell stories of myth and legend, |
| *Response:* | { "response": "No" } |
| *System Prompt:* | Does this prompt have multiple subjects or objects? If a pronoun relates to a noun in the sentence then it is considered a single subject. Respond with only a Yes or No. Only respond as a JSON object. For example, { "response": "Yes" }. |
| *Prompt:* | while the imposing stone structures rose up from the jungle like giants. |
| *Response:* | { "response": "No" } |
| *System Prompt:* | Does the given prompt imply a travel experience, whether it is in the past, present, or future? Respond with a Yes or a No. Only respond as a JSON object. For example, { "response": "Yes" }. |
| *Prompt:* | while the imposing stone structures rose up from the jungle like giants. |
| *Response:* | { "response": "No" } |
| *System Prompt:* | Does this prompt have multiple subjects or objects? If a pronoun relates to a noun in the sentence then it is considered a single subject. Respond with only a Yes or No. Only respond as a JSON object. For example, { "response": "Yes" }. |
| *Prompt:* | As I explored the ancient city, |
| *Response:* | { "response": "No" } |
| *System Prompt:* | Does the given prompt imply a travel experience, whether it is in the past, present, or future? Respond with a Yes or a No. Only respond as a JSON object. For example, { "response": "Yes" }. |
| *Prompt:* | As I explored the ancient city, |
| *Response:* | { "response": "Yes" } |
| *System Prompt:* | Analyze the sentence and identify all SVO phrases. Replace any pronouns (e.g. "he" or "she" or "they" or "I" or "you") with proper nouns. Respond with ONLY a JSON object with only three fields: subject, object, and action. For example, if I say "Ben is so happy - he bought a boat and is going sailing", you should reply with { "subject": "Ben", "object": "a boat", "action": "bought" } |
| *Prompt:* | As I explored the ancient city, |
| *Response:* | { "subject": "I", "object": "the ancient city", "action": "explored" } |
| *System Prompt:* | Does this prompt have multiple subjects or objects? If a pronoun relates to a noun in the sentence then it is considered a single subject. Respond with only a Yes or No. Only respond as a JSON object. For example, { "response": "Yes" }. |
| *Prompt:* | I felt a sense of connection to the people who had once called this place home - their history and culture still lingered in every stone. |
| *Response:* | { "response": "No" } |
| *System Prompt:* | Does the given prompt imply a travel experience, whether it is in the past, present, or future? Respond with a Yes or a No. Only respond as a JSON object. For example, { "response": "Yes" }. |
| *Prompt:* | I felt a sense of connection to the people who had once called this place home - their history and culture still lingered in every stone. |
| *Response:* | { "response": "Yes" } |
| *System Prompt:* | Analyze the sentence and identify all SVO phrases. Replace any pronouns (e.g. "he" or "she" or "they" or "I" or "you") with proper nouns. Respond with ONLY a JSON object with only three fields: subject, object, and action. For example, if I say "Ben is so happy - he bought a boat and is going sailing", you should reply with { "subject": "Ben", "object": "a boat", "action": "bought" } |
| *Prompt:* | I felt a sense of connection to the people who had once called this place home - their history and culture still lingered in every stone. |
| *Response:* | {<br>  "subject": "the people",<br>  "object": "their history and culture",<br>  "action": "lingered"<br>} |

**Listing A19.** Test 19.
Start: 9:09:05 AM      End: 9:09:25 AM

| | |
|---|---|
| *System Prompt:* | Is the following input multiple independent clauses? Respond with only a Yes or No. Only respond as a JSON object. For example, "Dillan ate an apple. He thought it tasted good. " { "response": "Yes" }. |
| *Prompt:* | As I sailed through the crystal-clear waters of the Aegean Sea, I felt a sense of freedom and joy that I had never experienced before. The picturesque villages and secluded beaches of the Greek Islands seemed to unfold like a treasure map before me - each new cove or hidden harbor revealed another breathtaking view or secret spot. As we anchored in a secluded bay for the night, I felt grateful for this chance to explore one of the most beautiful regions on earth. |
| *Response:* | { "response": "Yes" } |
| *System Prompt:* | Split this paragraph into distinct sentences. Only respond as a JSON array For example if given the text "This is one sentence. This is another sentence", respond with [ "This is one sentence.", "This is another sentence." ]. |
| *Prompt:* | As I sailed through the crystal-clear waters of the Aegean Sea, I felt a sense of freedom and joy that I had never experienced before. The picturesque villages and secluded beaches of the Greek |
| | Islands seemed to unfold like a treasure map before me - each new cove or hidden harbor revealed another breathtaking view or secret spot. As we anchored in a secluded bay for the night, I felt grateful for this chance to explore one of the most beautiful regions on earth. |
| *Response:* | [<br>  "As I sailed through the crystal-clear waters of the Aegean Sea, I felt a sense of freedom and joy that I had never experienced before.",<br>  "The picturesque villages and secluded beaches of the Greek Islands seemed to unfold like a treasure map before me - each new cove or hidden harbor revealed another breathtaking view or secret spot.",<br>  "As we anchored in a secluded bay for the night, I felt grateful for this chance to explore one of the most beautiful regions on earth."<br>] |
| *System Prompt:* | Does this prompt have multiple subjects or objects? If a pronoun relates to a noun in the sentence then it is considered a single subject. Respond with only a Yes or No. Only respond as a JSON object. For example, { "response": "Yes" }. |
| *Prompt:* | As I sailed through the crystal-clear waters of the Aegean Sea, I felt a sense of freedom and joy that I had never experienced before. |
| *Response:* | { "response": "No" } |
| *System Prompt:* | Does the given prompt imply a travel experience, whether it is in the past, present, or future? Respond with a Yes or a No. Only respond as a JSON object. For example, { "response": "Yes" }. |
| *Prompt:* | As I sailed through the crystal-clear waters of the Aegean Sea, I felt a sense of freedom and joy that I had never experienced before. |
| *Response:* | { "response": "Yes" } |
| *System Prompt:* | Analyze the sentence and identify all SVO phrases. Replace any pronouns (e.g. "he" or "she" or "they" or "I" or "you") with proper nouns. Respond with ONLY a JSON object with only three fields: subject, object, and action. For example, if I say "Ben is so happy - he bought a boat and is going sailing", you should reply with { "subject": "Ben", "object": "a boat", "action": "bought" } |
| *Prompt:* | As I sailed through the crystal-clear waters of the Aegean Sea, I felt a sense of freedom and joy that I had never experienced before. |
| *Response:* | {<br>  "subject": "I",<br>  "object": "freedom and joy",<br>  "action": "felt"<br>} |
| *System Prompt:* | Does this prompt have multiple subjects or objects? If a pronoun relates to a noun in the sentence then it is considered a single subject. Respond with only a Yes or No. Only respond as a JSON object. For example, { "response": "Yes" }. |
| *Prompt:* | The picturesque villages and secluded beaches of the Greek Islands seemed to unfold like a treasure map before me - each new cove or hidden harbor revealed another breathtaking view or secret spot. |
| *Response:* | { "response": "No" } |
| *System Prompt:* | Does the given prompt imply a travel experience, whether it is in the past, present, or future? Respond with a Yes or a No. Only respond as a JSON object. For example, { "response": "Yes" }. |
| *Prompt:* | The picturesque villages and secluded beaches of the Greek Islands seemed to unfold like a treasure map before me - each new cove or hidden harbor revealed another breathtaking view or secret spot. |
| *Response:* | { "response": "Yes" } |
| *System Prompt:* | Analyze the sentence and identify all SVO phrases. Replace any pronouns (e.g. "he" or "she" or "they" or "I" or "you") with proper nouns. Respond with ONLY a JSON object with only three fields: subject, object, and action. For example, if I say "Ben is so happy - he bought a boat and is |
| *Prompt:* | going sailing", you should reply with { "subject": "Ben", "object": "a boat", "action": "bought" }<br>The picturesque villages and secluded beaches of the Greek Islands seemed to unfold like a treasure map before me - each new cove or hidden harbor revealed another breathtaking view or secret spot. |
| *Response:* | {<br>  "subject": "the Greek Islands",<br>  "object": "a treasure map",<br>  "action": "unfold"<br>} |
| *System Prompt:* | Does this prompt have multiple subjects or objects? If a pronoun relates to a noun in the sentence then it is considered a single subject. Respond with only a Yes or No. Only respond as a JSON object. For example, { "response": "Yes" }. |
| *Prompt:* | As we anchored in a secluded bay for the night, I felt grateful for this chance to explore one of the most beautiful regions on earth. |
| *Response:* | { "response": "No" } |
| *System Prompt:* | Does the given prompt imply a travel experience, whether it is in the past, present, or future? Respond with a Yes or a No. Only respond as a JSON object. For example, { "response": "Yes" }. |
| *Prompt:* | As we anchored in a secluded bay for the night, I felt grateful for this chance to explore one of the most beautiful regions on earth. |
| *Response:* | { "response": "Yes" } |
| *System Prompt:* | Analyze the sentence and identify all SVO phrases. Replace any pronouns (e.g. "he" or "she" or "they" or "I" or "you") with proper nouns. Respond with ONLY a JSON object with only three fields: subject, object, and action. For example, if I say "Ben is so happy - he bought a boat and is going sailing", you should reply with { "subject": "Ben", "object": "a boat", "action": "bought" } |
| *Prompt:* | As we anchored in a secluded bay for the night, I felt grateful for this chance to explore one of the most beautiful regions on earth. |
| *Response:* | {<br>  "subject": "I",<br>  "object": "this chance",<br>  "action": "felt"<br>} |

**Listing A20.** Test 20.
Start: 9:11:03 AM        End: 9:11:22 AM

| | |
|---|---|
| *System Prompt:* | Is the following input multiple independent clauses? Respond with only a Yes or No. Only respond as a JSON object. For example, "Dillan ate an apple. He thought it tasted good. " { "response": "Yes" }. |
| *Prompt:* | As I stood before the majestic Grand Canyon at sunset, I felt a sense of awe and wonder wash over me. The rust-red rock formations seemed to glow with an otherworldly light in the fading light of day - the sheer scale and beauty of this incredible natural wonder left me speechless. As I walked along the rim trail, taking in the breathtaking views and listening to the stories of the Native American tribes who had once called this place home, I felt a sense of connection to the land itself. |
| *Response:* | { "response": "Yes" } |
| *System Prompt:* | Split this paragraph into distinct sentences. Only respond as a JSON array For example if given the text "This is one sentence. This is another sentence", respond with [ "This is one sentence.", "This is another sentence." ]. |
| *Prompt:* | As I stood before the majestic Grand Canyon at sunset, I felt a sense of awe and wonder wash over me. The rust-red rock formations seemed to glow with an otherworldly light in the fading light of day - the sheer scale and beauty of this incredible natural wonder left me speechless. As I walked along the rim trail, taking in the breathtaking views and listening to the stories of the Native American tribes who had once called this place home, I felt a sense of connection to the land itself. |
| *Response:* | [<br>   "As I stood before the majestic Grand Canyon at sunset, I felt a sense of awe and wonder wash over me.",<br>   "The rust-red rock formations seemed to glow with an otherworldly light in the fading light of day - the sheer scale and beauty of this incredible natural wonder left me speechless.",<br>   "As I walked along the rim trail, taking in the breathtaking views and listening to the stories of the Native American tribes who had once called this place home, I felt a sense of connection to the land itself."<br>] |
| *System Prompt:* | Does this prompt have multiple subjects or objects? If a pronoun relates to a noun in the sentence then it is considered a single subject. Respond with only a Yes or No. Only respond as a JSON object. For example, { "response": "Yes" }. |
| *Prompt:* | As I stood before the majestic Grand Canyon at sunset, I felt a sense of awe and wonder wash over me. |
| *Response:* | { "response": "No" } |
| *System Prompt:* | Does the given prompt imply a travel experience, whether it is in the past, present, or future? Respond with a Yes or a No. Only respond as a JSON object. For example, { "response": "Yes" }. |
| *Prompt:* | As I stood before the majestic Grand Canyon at sunset, I felt a sense of awe and wonder wash over me. |
| *Response:* | { "response": "Yes" } |
| *System Prompt:* | Analyze the sentence and identify all SVO phrases. Replace any pronouns (e.g. "he" or "she" or "they" or "I" or "you") with proper nouns. Respond with ONLY a JSON object with only three fields: subject, object, and action. For example, if I say "Ben is so happy - he bought a boat and is going sailing", you should reply with { "subject": "Ben", "object": "a boat", "action": "bought" } |
| *Prompt:* | As I stood before the majestic Grand Canyon at sunset, I felt a sense of awe and wonder wash over me. |
| *Response:* | {<br>  "subject": "I",<br>  "object": "awe and wonder",<br>  "action": "felt"<br>} |
| *System Prompt:* | Does this prompt have multiple subjects or objects? If a pronoun relates to a noun in the sentence then it is considered a single subject. Respond with only a Yes or No. Only respond as a JSON object. For example, { "response": "Yes" }. |
| *Prompt:* | The rust-red rock formations seemed to glow with an otherworldly light in the fading light of day - the sheer scale and beauty of this incredible natural wonder left me speechless. |
| *Response:* | { "response": "No" } |
| *System Prompt:* | Does the given prompt imply a travel experience, whether it is in the past, present, or future? Respond with a Yes or a No. Only respond as a JSON object. For example, { "response": "Yes" }. |
| *Prompt:* | The rust-red rock formations seemed to glow with an otherworldly light in the fading light of day - the sheer scale and beauty of this incredible natural wonder left me speechless. |
| *Response:* | { "response": "Yes" } |
| *System Prompt:* | Analyze the sentence and identify all SVO phrases. Replace any pronouns (e.g. "he" or "she" or "they" or "I" or "you") with proper nouns. Respond with ONLY a JSON object with only three fields: subject, object, and action. For example, if I say "Ben is so happy - he bought a boat and is going sailing", you should reply with { "subject": "Ben", "object": "a boat", "action": "bought" } |
| *Prompt:* | The rust-red rock formations seemed to glow with an otherworldly light in the fading light of day - the sheer scale and beauty of this incredible natural wonder left me speechless. |
| *Response:* | {<br>  "subject": "the rust-red rock formations", |

```
      "object": "me",
      "action": "left"
    }
```

*System Prompt:* Does this prompt have multiple subjects or objects? If a pronoun relates to a noun in the sentence then it is considered a single subject. Respond with only a Yes or No. Only respond as a JSON object. For example, { "response": "Yes" }.
*Prompt:* As I walked along the rim trail, taking in the breathtaking views and listening to the stories of the Native American tribes who had once called this place home, I felt a sense of connection to the land itself.
*Response:* { "response": "No" }

*System Prompt:* Does the given prompt imply a travel experience, whether it is in the past, present, or future? Respond with a Yes or a No. Only respond as a JSON object. For example, { "response": "Yes" }.
*Prompt:* As I walked along the rim trail, taking in the breathtaking views and listening to the stories of the Native American tribes who had once called this place home, I felt a sense of connection to the land itself.
*Response:* { "response": "Yes" }

*System Prompt:* Analyze the sentence and identify all SVO phrases. Replace any pronouns (e.g. "he" or "she" or "they" or "I" or "you") with proper nouns. Respond with ONLY a JSON object with only three fields: subject, object, and action. For example, if I say "Ben is so happy - he bought a boat and is going sailing", you should reply with { "subject": "Ben", "object": "a boat", "action": "bought" }
*Prompt:* As I walked along the rim trail, taking in the breathtaking views and listening to the stories of the Native American tribes who had once called this place home, I felt a sense of connection to the land itself.
*Response:*
```
    {
      "subject": "I",
      "object": "the land",
      "action": "felt"
    }
```

**Listing A21.** Test 21.
Start: 9:12:35 AM          End: 9:12:55 AM

*System Prompt:* Is the following input multiple independent clauses? Respond with only a Yes or No. Only respond as a JSON object. For example, "Dillan ate an apple. He thought it tasted good. " { "response": "Yes" }.
*Prompt:* As I rode through the picturesque villages and mountain passes of the Swiss Alps on a vintage train, I felt a sense of nostalgia and wonder wash over me. The snow-capped peaks and turquoise lakes seemed to unfold like a postcard before me - each new station or village revealed another breathtaking view or hidden gem. As we chugged along the tracks, I felt grateful for this chance to experience one of the most scenic train rides in the world.
*Response:* { "response": "Yes" }

*System Prompt:* Split this paragraph into distinct sentences. Only respond as a JSON array For example if given the text "This is one sentence. This is another sentence", respond with [ "This is one sentence.", "This is another sentence." ].
*Prompt:* As I rode through the picturesque villages and mountain passes of the Swiss Alps on a vintage train, I felt a sense of nostalgia and wonder wash over me. The snow-capped peaks and turquoise lakes seemed to unfold like a postcard before me - each new station or village revealed another breathtaking view or hidden gem. As we chugged along the tracks, I felt grateful for this chance to experience one of the most scenic train rides in the world.
*Response:*
```
    [
       "As I rode through the picturesque villages and mountain passes of the Swiss Alps on a vintage train, I felt a sense of nostalgia and wonder wash over me.",
```

```
        "The snow-capped peaks and turquoise lakes seemed to unfold like a postcard before me - each
    new station or village revealed another breathtaking view or hidden gem.",
        "As we chugged along the tracks, I felt grateful for this chance to experience one of the most
    scenic train rides in the world."
    ]
```

| | |
|---|---|
| *System Prompt:* | Does this prompt have multiple subjects or objects? If a pronoun relates to a noun in the sentence then it is considered a single subject. Respond with only a Yes or No. Only respond as a JSON object. For example, { "response": "Yes" }. |
| *Prompt:* | As I rode through the picturesque villages and mountain passes of the Swiss Alps on a vintage train, I felt a sense of nostalgia and wonder wash over me. |
| *Response:* | { "response": "No" } |
| *System Prompt:* | Does the given prompt imply a travel experience, whether it is in the past, present, or future? Respond with a Yes or a No. Only respond as a JSON object. For example, { "response": "Yes" }. |
| *Prompt:* | As I rode through the picturesque villages and mountain passes of the Swiss Alps on a vintage train, I felt a sense of nostalgia and wonder wash over me. |
| *Response:* | { "response": "Yes" } |
| *System Prompt:* | Analyze the sentence and identify all SVO phrases. Replace any pronouns (e.g. "he" or "she" or "they" or "I" or "you") with proper nouns. Respond with ONLY a JSON object with only three fields: subject, object, and action. For example, if I say "Ben is so happy - he bought a boat and is going sailing", you should reply with { "subject": "Ben", "object": "a boat", "action": "bought" } |
| *Prompt:* | As I rode through the picturesque villages and mountain passes of the Swiss Alps on a vintage train, I felt a sense of nostalgia and wonder wash over me. |
| *Response:* | {<br>  "subject": "I",<br>  "object": "a sense of nostalgia and wonder",<br>  "action": "wash"<br>} |
| *System Prompt:* | Does this prompt have multiple subjects or objects? If a pronoun relates to a noun in the sentence then it is considered a single subject. Respond with only a Yes or No. Only respond as a JSON object. For example, { "response": "Yes" }. |
| *Prompt:* | The snow-capped peaks and turquoise lakes seemed to unfold like a postcard before me - each new station or village revealed another breathtaking view or hidden gem. |
| *Response:* | { "response": "No" } |
| *System Prompt:* | Does the given prompt imply a travel experience, whether it is in the past, present, or future? Respond with a Yes or a No. Only respond as a JSON object. For example, { "response": "Yes" }. |
| *Prompt:* | The snow-capped peaks and turquoise lakes seemed to unfold like a postcard before me - each new station or village revealed another breathtaking view or hidden gem. |
| *Response:* | { "response": "Yes" } |
| *System Prompt:* | Analyze the sentence and identify all SVO phrases. Replace any pronouns (e.g. "he" or "she" or "they" or "I" or "you") with proper nouns. Respond with ONLY a JSON object with only three fields: subject, object, and action. For example, if I say "Ben is so happy - he bought a boat and is going sailing", you should reply with { "subject": "Ben", "object": "a boat", "action": "bought" } |
| *Prompt:* | The snow-capped peaks and turquoise lakes seemed to unfold like a postcard before me - each new station or village revealed another breathtaking view or hidden gem. |
| *Response:* | {<br>  "subject": "the snow-capped peaks and turquoise lakes",<br>  "object": "a postcard",<br>  "action": "seemed"<br>} |
| *System Prompt:* | Does this prompt have multiple subjects or objects? If a pronoun relates to a noun in the sentence then it is considered a single subject. Respond with only a Yes or No. Only respond as a JSON object. For example, { "response": "Yes" }. |
| *Prompt:* | As we chugged along the tracks, I felt grateful for this chance to experience one of the most scenic train rides in the world. |
| *Response:* | { "response": "No" } |
| | |
| *System Prompt:* | Does the given prompt imply a travel experience, whether it is in the past, present, or future? Respond with a Yes or a No. Only respond as a JSON object. For example, { "response": "Yes" }. |
| *Prompt:* | As we chugged along the tracks, I felt grateful for this chance to experience one of the most scenic train rides in the world. |
| *Response:* | { "response": "Yes" } |
| | |
| *System Prompt:* | Analyze the sentence and identify all SVO phrases. Replace any pronouns (e.g. "he" or "she" or "they" or "I" or "you") with proper nouns. Respond with ONLY a JSON object with only three fields: subject, object, and action. For example, if I say "Ben is so happy - he bought a boat and is going sailing", you should reply with { "subject": "Ben", "object": "a boat", "action": "bought" } |
| *Prompt:* | As we chugged along the tracks, I felt grateful for this chance to experience one of the most scenic train rides in the world. |
| *Response:* | {<br>  "subject": "I",<br>  "object": "one of the most scenic train rides in the world",<br>  "action": "felt"<br>} |